\pdfoutput=1

\documentclass[11pt]{article}

\usepackage[preprint]{acl_latex}

\usepackage{times}
\usepackage{latexsym}

\usepackage[T1]{fontenc}

\usepackage[utf8]{inputenc}

\usepackage{microtype}

\usepackage{inconsolata}

\usepackage{graphicx}
\usepackage{subcaption}

\usepackage{algorithm}
\usepackage{algorithmic}

\usepackage{xcolor}
\usepackage{colortbl}
\definecolor{darkred}{RGB}{200,20,20}

\usepackage{tabularx}
\usepackage{adjustbox}
\usepackage{booktabs}
\usepackage{multirow}

\usepackage{enumitem}
\usepackage{setspace}

\usepackage{newfloat}
 \usepackage{stfloats}
\usepackage{listings}

\usepackage{placeins}

\title{Tokenization Matters! Degrading Large Language Models through Challenging Their Tokenization}

\author{
  Dixuan Wang\textsuperscript{\rm 1},
  Yanda Li\textsuperscript{\rm 1},
  Junyuan Jiang\textsuperscript{\rm 2},
  Zepeng Ding\textsuperscript{\rm 1},
  Ziqin Luo\textsuperscript{\rm 1},
  Guochao Jiang\textsuperscript{\rm 1},\\
  \textbf{Jiaqing Liang}\textsuperscript{\rm 1}\thanks{Corresponding Authors},
  \textbf{Deqing Yang}\textsuperscript{\rm 1}\footnotemark[1]
  \\
  School of Data Science, Fudan University, Shanghai, China \textsuperscript{\rm 1}\\
  School of Management, Fudan University, Shanghai, China \textsuperscript{\rm 2}\\
  \texttt{\{dxwang23,ydli22,jiangjy21,zqluo22,gcjiang22\}@m.fudan.edu.cn}, \\
  \texttt{\{dingzepeng,liangjiaqing,yangdeqing\}@fudan.edu.cn}
}

\begin{document}
\maketitle
\begin{abstract}
  Large Language Models (LLMs) have shown remarkable capabilities in language understanding and generation. Nonetheless, it was also witnessed that LLMs tend to produce inaccurate responses to specific queries. This deficiency can be traced to the tokenization step LLMs must undergo, which is an inevitable limitation inherent to all LLMs. In fact, incorrect tokenization is the critical point that hinders LLMs in understanding the input precisely, thus leading to unsatisfactory output. 
  This defect is more obvious in Chinese scenarios. To demonstrate this flaw of LLMs, we construct an adversarial dataset, named as \textbf{ADT (Adversarial Dataset for Tokenizer)}, which draws upon the vocabularies of various open-source LLMs to challenge LLMs' tokenization. ADT consists of two subsets: the manually constructed ADT-Human and the automatically generated ADT-Auto. Our empirical results reveal that our ADT is highly effective on challenging the tokenization of leading LLMs, including GPT-4o, Llama-3, Deepseek-R1 and so on, thus degrading these LLMs' capabilities. Moreover, our method of automatic data generation has been proven efficient and robust, which can be applied to any open-source LLMs. 
  In this paper, we substantially investigate LLMs' vulnerability in terms of challenging their token segmentation, which will shed light on the subsequent research of improving LLMs' capabilities through optimizing their tokenization process and algorithms.
\end{abstract}

\section{Introduction}
\label{Introduction}



In the last two years, Large Language Models (LLMs) have demonstrated remarkable capabilities in many tasks of artificial intelligence (AI), including natural language generation \cite{hoffmann2022training, nijkamp2023codegen, zeng2023glm130b}, knowledge utilization~\cite{chowdhery2022palm, izacard2022atlas}, and complex reasoning~\cite{wei2023chainofthought, zhou2023leasttomost, kojima2023large, taylor2022galactica}. 
Given these capabilities, LLMs have been effectively employed by various application domains, such as healthcare~\cite{tang2023does, li2023chatdoctor, jeblick2022chatgpt}, education~\cite{susnjak2022chatgpt, bordt2023chatgpt, Malinka_2023}, law~\cite{yu2022legal, nay2023law} and so on.

\begin{figure}
    \centering
    \includegraphics[width=1\linewidth]{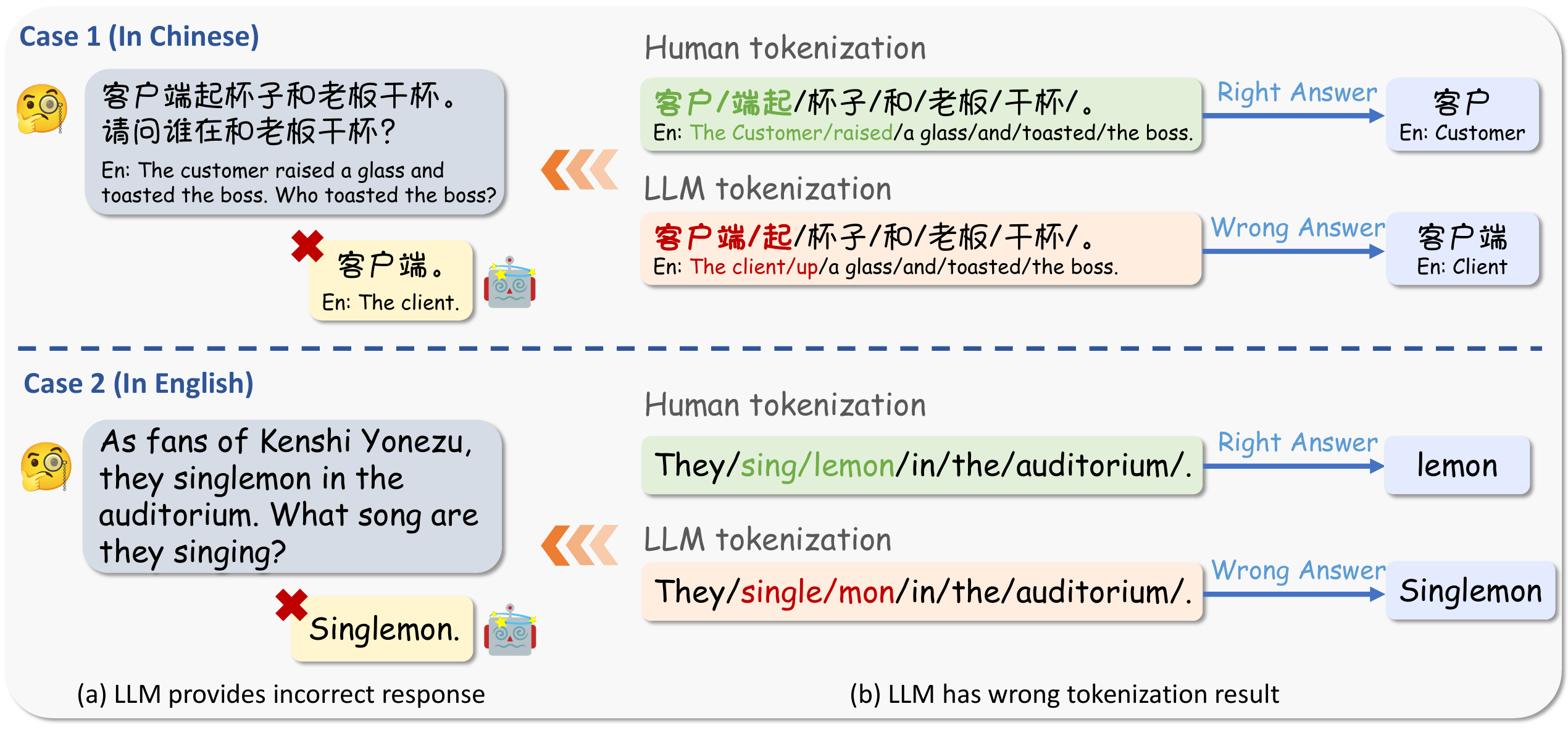}
    \caption{Two instances of LLM generating incorrect response due to incorrect tokenization. Case 1 is a Chinese input instance, of which the English translation is noted below according to its correct tokenization. In Case 2, a space is omitted between `sing' and `lemon', causing the LLM's incorrect tokenization, which is detailed in Section \ref{Construction}.}
    \label{fig:example_mix}
\end{figure}

Nonetheless, LLMs' disadvantages have also been witnessed, including hallucination~\cite{openai2023gpt, bang2023multitask, lin2022truthfulqa}, knowledge recency~\cite{dai2022knowledge, Kernbach_2022}, and so on. 
Particularly, we observed that for some specific queries, LLMs often produce unsatisfactory responses with words that are nonsensical, as illustrated by the two instances in Figure~\ref{fig:example_mix}. 
Through checking the LLM's tokenization results for input sentences, we found that it is misaligned with human's correct comprehension for the sentences. 
Notably, our empirical studies found that this flaw not only exists in some specific LLMs, but also is a universal issue across many mainstream LLMs. 
We have evaluated several prominent open-source and closed-source LLMs, including Chatglm3 \cite{zeng2023glm130b}, 
Qwen2.5-max~\cite{qwen25}, 
Deepseek-R1 \cite{deepseekai2025deepseekr1incentivizingreasoningcapability}, and GPT-4o~\cite{openai2024gpt4ocard}. 
Our experiment results reveal that regardless of LLMs' scales or their claimed capabilities, they inevitably generate incorrect or entirely nonsensical responses for some specific inputs when their tokenization results for the input sentences are obviously wrong. 
Consequently, we believe that LLMs' tokenization errors prevent them from accurately understanding the input text, leading to their incorrect responses. 

As we know, LLMs' tokenization flaws stem from the algorithms of their tokenizers, most of which are based on subword-level vocabularies. 
The popular tokenization algorithms include Byte-Pair Encoding (BPE) \cite{sennrich2016neural}, WordPiece \cite{Schuster_Nakajima_2012}, and Unigram \cite{kudo2018subword}. 
However, no vocabulary can perfectly cover all possible ways of various expressions in the inputs. 
The algorithms may potentially generate unsatisfactory results which are not fully aligned with the true intention of users' input. 
Unfortunately, in cases of tokenization errors, all subsequent optimization operations for LLMs cannot completely solve this underlying problems caused by their tokenization algorithms.

In the domain of natural language processing (NLP), the existing studies related to tokenization primarily focus on refining or optimizing various tokenization algorithms. 
Meanwhile, the discussions on LLMs' vulnerability including attack or challenge techniques, have been more concerned with the security of LLMs. 
For LLMs, in terms of the challenges posed by tokenization deficiencies, Sander and Max~\cite{land2024fishing} have discussed this issue from the perspective of under-trained tokens in LLMs. 
It is worth noting that, to the best of our knowledge, there has been no research specifically focusing on the unsatisfactory token segmentation of LLMs, particularly in Chinese scenarios, which is indeed a critical concern causing LLMs' vulnerability.

In this paper, we focus on the critical flaw in LLMs' tokenization process, and try to reveal the relationship between LLMs' unsatisfactory tokenization and their inaccurate responses for some specific queries. 
To this end, for the first time, we construct a dataset, namely \textbf{ADT (Adversarial Dataset for Tokenizer)}, to challenge the tokenization of various LLMs. 
ADT dataset consists of two subsets: the manually constructed \textbf{ADT-Human} and the automatically generated \textbf{ADT-Auto}. 
At first, we export the vocabularies from multiple mainstream LLMs, based on which ADT-Human is constructed. 
Our experiment results demonstrate that ADT-Human can effectively challenge LLMs' tokenization, leading to their completely incorrect responses. 
Furthermore, we also develop a framework for automatically generating adversarial data to construct dataset more efficiently. 
Initially, we export LLM's vocabulary and identify the trap words that can influence the model's performance. 
By inputting these trap words into GPT-4~\cite{openai2023gpt} with prompt, we leverage its capability to get available instances, which are then inspected manually to ensure quality. 
With minimal human effort, we successfully construct ADT-Auto with 231 instances, validating the effectiveness of our framework and highlighting the inevitable flaws in LLMs' tokenization once again.

In summary, the contributions of this paper are as follows:
\begin{enumerate}[topsep=3pt, leftmargin=12pt]
    \setlength{\itemsep}{2pt} 
    \setlength{\parsep}{0pt}  
    \setlength{\parskip}{0pt} 
    \item 
    We investigate LLMs' vulnerability for some special inputs in terms of challenging their token segmentation, 
    which provides a new perspective of studying LLMs' disadvantages.
    \item We propose an effective framework to construct a new dataset ADT, which can be used to challenge various LLMs' tokenization, exposing their vulnerability for specific queries.
    \item Our experiment results obviously reveal the relationship between LLMs' unsatisfactory tokenization and inaccurate responses, which can shed light on subsequent work of improving LLMs through optimizing tokenization. 
    
\end{enumerate}

\section{Related Work}
\label{Related Work}

\paragraph{Algorithms for tokenization.}
Tokenization is a basic but crucial step in NLP. 
Currently, the mainstream tokenization approachs are based on subwords, including Byte Pair Encoding (BPE), WordPiece and Unigram.
BPE~\cite{sennrich2016neural} forms the vocabulary starting from character-level tokens, merging token pairs and intercalating them into vocabulary. The merging rule is to select adjacent pairs with the highest word frequency. LLMs including GPT-3~\cite{brown2020language}, RoBERTa~\cite{liu2019roberta}, 
and Llama2~\cite{touvron2023llama} 
are based on BPE.
WordPiece~\cite{Schuster_Nakajima_2012} is similar to BPE, but it differs in the strategy of merging pairs with reference to mutual information rather than frequency. LLMs built upon WordPiece include BERT~\cite{devlin2019bert}, DistilBERT~\cite{sanh2020distilbert}, and Electra~\cite{clark2020electra}. 
Different from these two algorithms, Unigram~\cite{kudo2018subword} starts with a large vocabulary and gradually trims it down to a smaller one. It measures the importance of subwords by calculating loss associated with the removal of each subword, ultimately retaining those exhibit high importance.
LLMs utilizing Unigram include AlBERT~\cite{lan2020albert} and mBART~\cite{liu2020multilingual}. 
Besides, many tokenization tools integrating these algorithms have been developed, such as Google's SentencePiece~\cite{kudo2018sentencepiece} and OpenAI's tiktoken\footnote{\url{https://github.com/openai/tiktoken}}, which simplify the process of LLMs tokenization greatly. 
In recent years, it has been investigated how the input segmentation of pre-trained language models (PLMs) affects the interpretations of derivationally complex English words~\cite{hofmann-etal-2021-superbizarre}. Some scholars have proposed FLOTA~\cite{hofmann-etal-2022-embarrassingly}, a simple yet effective method to improve PLMs' tokenization of English words.
However, there has been no work concerning the shared risks of LLMs in terms of inaccurate tokenization result.
We focus on this issue, discussing the underlying risks of LLMs' tokenization.

\paragraph{Attack techniques in LLMs.}
With the growing prominence of LLMs, the security and vulnerability of these models have attracted significant attention, and even advanced LLMs like GPT-4 are no exception. A surge of research in this field is underway, with researchers launching attacks on LLMs from various aspects~\cite{esmradi2023comprehensive, chowdhury2024breaking} including but not limited to, Prompt Injection~\cite{choi2022prompt}, Model Theft~\cite{krishna2020thieves}, Data Reconstruction~\cite{carlini2021extracting}, Data Poisoning\cite{wallace2021concealed, xu2024instructions}, and Member Inference Attack~\cite{7958568}. 
For instance, Prompt Injection Attack refers to a scenario where an attacker crafts malicious prompts to deceive language models into generating outputs inconsistent with their training data and anticipated functionality. 
Threat actors aim at information gathering, fraud, intrusion, content manipulation, and availability attacks \cite{choi2022prompt}. 
Carlini, Nicholas, et al~\cite{carlini2021extracting} executed the Data Reconstruction attack on GPT-2, extracting personal identity information, code, and UUIDs. 
Data Poisoning pertains to the deliberate introduction of corrupted or malicious data into the training dataset to manipulate the model's behavior. 
Diverging from existing works, our research innovatively suggests attacking the capabilities of LLMs from the perspective of tokenization.

\section{Methodology of Dataset Construction}
\label{Method}

In this section, we describe the process of constructing \textbf{ADT (Adversarial Dataset for Tokenizer)} in detail, which is used to challenge LLMs' tokenization and thus reveal LLMs' vulnerability. 
ADT contains two subsets, manually constructed ADT-Human and automatically generated ADT-Auto.

\subsection{Vocabulary Export}
\label{Exporting vocabularies}
In fact, an instance in our dataset comprises one sentence containing the token (word) that could challenge LLMs' tokenization and one question related to the sentence. 
To inspect whether LLMs can accurately recognize these challenging tokens within various contexts based on their memories, the tokens should come from vocabularies of LLMs. 
Therefore, the first step of dataset construction is to export vocabularies of LLMs. 
Given that Chinese is more complex and challenging than English in terms of tokenization, as detailed in Section \ref{Construction}, we pay more attention to the issue in Chinese data.

Specifically, we select five widely-used open-source LLMs which are trained on extensive Chinese corpus, to export their Chinese vocabularies, including Chatglm3-6B~\cite{zeng2023glm130b}, Baichuan2-13B-Chat~\cite{yang2023baichuan}, Yi-34B-Chat~\cite{ai2024yi}, Qwen-7B-Chat~\cite{bai2023qwen}, and Qwen1.5-72B-Chat~\cite{bai2023qwen}. 
Besides, we export the English vocabularies from three English-based LLMs, i.e., Llama-3-8B-Instruct~\cite{touvron2023llama}, Llama-3-70B-Instruct~\cite{touvron2023llama}, and Mixtral-8x7B-Instruct-v0.1~\cite{jiang2024mixtral}. 
Notably, the Chinese vocabulary of Qwen-7B-Chat and Qwen1.5-72B-Chat is the same, so is the English vocabulary of Llama-3-8B-Instruct and Llama-3-70B-Instruct. 

\begin{table}
  \centering
  \begin{adjustbox}{max width=\columnwidth}
  \begin{tabular}{llcc}
    \toprule
     \multirow{2}{*}{\textbf{Language}} & \multirow{2}{*}{\textbf{Model}} & \multirow{2}{*}{\textbf{Vocabulary Size}} & \textbf{Vocabulary Size}\\
     & & & \textbf{of Specific Language} \\
    \midrule
    \multirow{5}{*}{\textbf{Chinese}} & Chatglm3-6B & 64,789 & 30,922\\
     & Baichuan2-13B-Chat & 125,696 & 70,394 \\
     & Yi-34B-Chat & 64,000 & 21,353 \\
     & Qwen-7B-Chat & 151,851 & 24,953 \\
     & Qwen1.5-72B-Chat & 151,646 & 24,953 \\
    \midrule
    \multirow{3}{*}{\textbf{English}} & Llama-3-8B-Instruct & 128,257 & 72,420 \\
     & Llama-3-70B-Instruct & 128,257 & 72,420 \\
     & Mixtral-8x7B-Instruct-v0.1 & 32,000 & 25,056 \\
    \bottomrule
  \end{tabular}
  \end{adjustbox}
  \caption{Vocabulary sizes of different LLMs.}
  \label{tab:size}
\end{table}

In the process of exporting vocabularies, a sequence of straightforward operations is considered. 
Initially, the tokenizer is decoded to obtain vocabulary, followed by the removal of leading and trailing spaces from each token. 
Notably, if SentencePiece is used for tokenization in training phase, some certain tokens may begin with a special token `\textunderscore' because SentencePiece treats the input text just as a sequence of Unicode characters, and whitespace is also handled as a normal symbol. 
To handle the whitespace as a basic token explicitly, SentencePiece first escapes the whitespace with a meta symbol `\textunderscore' (U+2581)~\cite{kudo2018sentencepiece}.
Consequently, when exporting vocabularies for models trained with SentencePiece, we replace `\textunderscore' with a blank character.
We summarize the exported vocabulary sizes of different LLMs in Table~\ref{tab:size}.

\subsection{ADT-Human Construction}
\label{Construction}

\begin{figure*}
    \centering
    \includegraphics[width=0.98\linewidth]{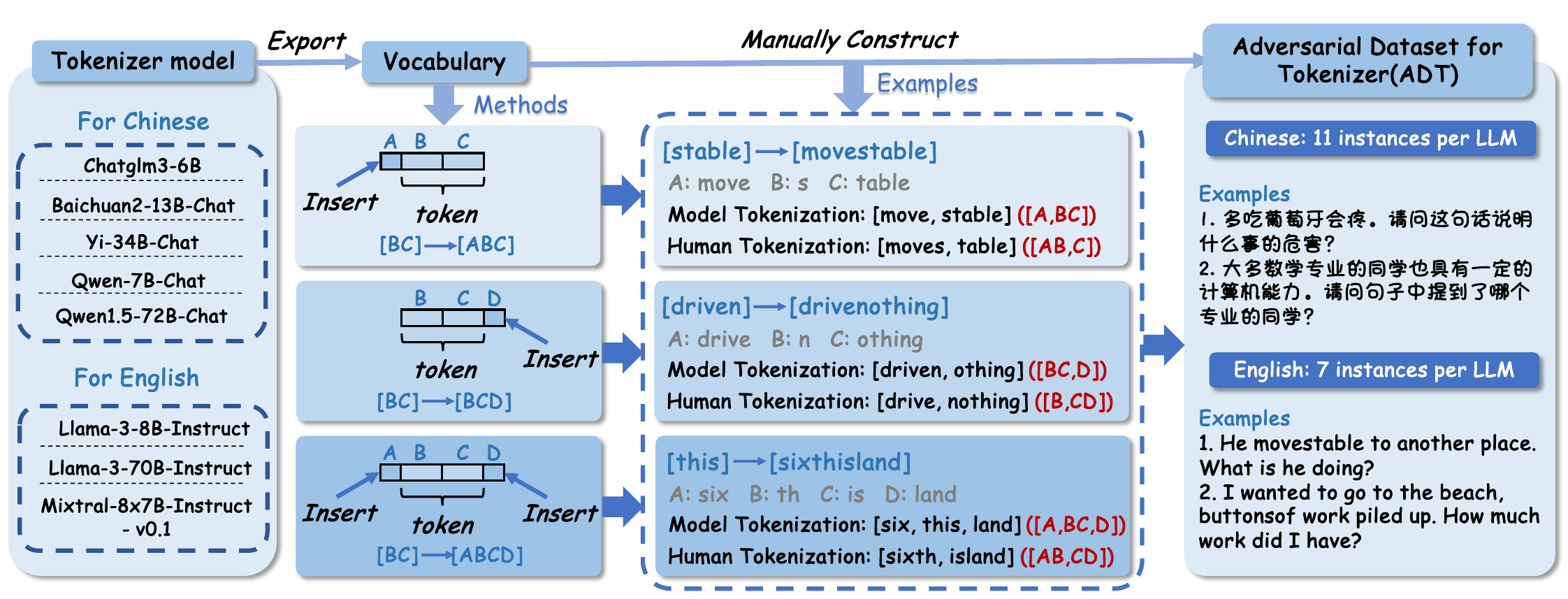}
    \caption{Our framework of constructing ADT-Human manually.}
    \label{fig:manually}
\end{figure*}

Based on exported vocabularies, we manually construct dataset ADT-Human, to challenge and evaluate the tokenization of different LLMs. 
The process of manual construction is depicted in Figure \ref{fig:manually}.

\begin{table*}
  \centering
  \begin{adjustbox}{max width=0.95\textwidth}
  \begin{tabular}{cccccc}
    \toprule
    {\textbf{Approach}} & & \textbf{Origin token} & \textbf{After insertion} & \textbf{Model tokenization} & \textbf{Human tokenization} \\
    & & & (Challenging span) & & \\
    \midrule
    \multirow{2}{*}{\textbf{Before}} & \textbf{Schema} & \texttt{\textcolor{cyan}{B}\textcolor{cyan!60!blue}{C}} & \texttt{\textcolor{orange}{A}\textcolor{cyan}{B}\textcolor{cyan!60!blue}{C}} & \texttt{[A, BC]} & \texttt{[AB, C]}  \\
    \multirow{2}{*} & \textbf{Example} & \textcolor{cyan}{s}\textcolor{cyan!60!blue}{table} & \textcolor{orange}{move}\textcolor{cyan}{s}\textcolor{cyan!60!blue}{table} & [move, stable] & [moves, table] \\
    \midrule
    \multirow{2}{*}{\textbf{After}} & \textbf{Schema} & \texttt{\textcolor{cyan}{A}\textcolor{cyan!60!blue}{B}} & \texttt{\textcolor{cyan}{A}\textcolor{cyan!60!blue}{B}\textcolor{orange}{C}} & \texttt{[AB, C]} & \texttt{[A, BC]} \\
    \multirow{2}{*} & \textbf{Example} & \textcolor{cyan}{drive}\textcolor{cyan!60!blue}{n} & \textcolor{cyan}{drive}\textcolor{cyan!60!blue}{n}\textcolor{orange}{othing} & [driven, othing] & [drive, nothing] \\
    \midrule
    \multirow{2}{*}{\textbf{Before \& After}} & \textbf{Schema} & \texttt{\textcolor{cyan}{B}\textcolor{cyan!60!blue}{C}} & \texttt{\textcolor{orange}{A}\textcolor{cyan}{B}\textcolor{cyan!60!blue}{C}\textcolor{orange}{D}} & \texttt{[A, BC, D]} & \texttt{[AB, CD]}  \\
    \multirow{2}{*} & \textbf{Example} & \textcolor{cyan}{th}\textcolor{cyan!60!blue}{is} & \textcolor{orange}{six}\textcolor{cyan}{th}\textcolor{cyan!60!blue}{is}\textcolor{orange}{land} & [six, this, land] & [sixth, island] \\
    \bottomrule
  \end{tabular}
  \end{adjustbox}
  \caption{Three approaches of generating challenging spans.}
  \label{tab:three_methods}
\end{table*}

Our purpose is to confirm the existence of challenges in tokenization, so for each LLM, a certain amount of data is constructed, which does not need to be large.
Specifically, we select eleven tokens from each Chinese vocabulary, and seven tokens for each English vocabulary. The main criterion for selecting tokens is that they should be easy to make sentences with.
According to the experimental results presented in Section \ref{experiment_human}, these data can effectively challenge the performance of LLMs, proving this issue deserves more attention. As for efficiently generating data in bulk, we design the framework for automatic generation in Section \ref{AutoGen}.

Then, for each selected token, we adopt one of the three approaches listed in Table~\ref{tab:three_methods}, to convert it into a \emph{challenging span} through inserting a special character span before or (and) after it. 
These challenging spans would disrupt the conventional tokenization process, thus confusing LLMs. 
The schemas and examples of three approaches are also shown in the second and third parts of Figure \ref{fig:manually}. 
The considerations in the three conversion approaches are introduced as follows.

\begin{enumerate}[topsep=3pt, leftmargin=12pt]
    \setlength{\itemsep}{2pt} 
    \setlength{\parsep}{0pt}  
    \setlength{\parskip}{0pt} 
    \item \textbf{Before}: A character span $s$ is inserted {before} origin token, causing the concatenation of $s$ and the token's prefix is just another token existing in vocabulary, as `move'+`s'$\rightarrow$`moves' in Table~\ref{tab:three_methods}.
    
    \item \textbf{After}: A character span $s$ is inserted {after} origin token, causing the concatenation of the token's suffix and $s$ is just another token existing in vocabulary, as `n'+`othing'$\rightarrow$`nothing' in Table~\ref{tab:three_methods}.
    
    \item \textbf{Before \& After}: Character spans $s_1$ and $s_2$ are inserted before and after origin token, respectively, causing the concatenation of $s_1$ and the token's prefix is just another token existing in vocabulary, meanwhile so is the concatenation of the token's suffix and $s_2$, as `six'+`th'$\rightarrow$`sixth' and `is'+`land'$\rightarrow$`island' in Table~\ref{tab:three_methods}.
\end{enumerate}

Next, for each challenging span $s_c$, we manually compose a corresponding instance as the data in ADT-Human. 
One instance consists of a sentence in which $s_c$ presents, and a corresponding question of which the answer comes from $s_c$. 
In Appendix~\ref{app:ADT}, we list all instances (along with their correct tokenizations) of ADT-Human.
As we can see, it is challenging for LLMs to understand the instances due to the presence of challenging spans.

It's worth noting that the tokenization difficulty of English is less than that of Chinese, since spaces are regularly used as delimiters to separate each word from others in English. 
Moreover, affixation is common in English word structure, enabling tokenizers to divide a single word into several sections, which can help avoid incorrect tokenization to some extent. 
Thus, during the manual construction of English instances, we deliberately exclude the spaces between some tokens to provoke challenges to tokenization process. 
This decision stems from the recognition that powerful models should possess robust abilities across various scenarios which can occur in real-world applications, including handling cases where spaces might be omitted in English text input~\cite{hofmann-etal-2022-embarrassingly}.


\subsection{ADT-Auto Generation}
\label{AutoGen}

\begin{figure*}
    \centering
    \includegraphics[width=0.96\linewidth]{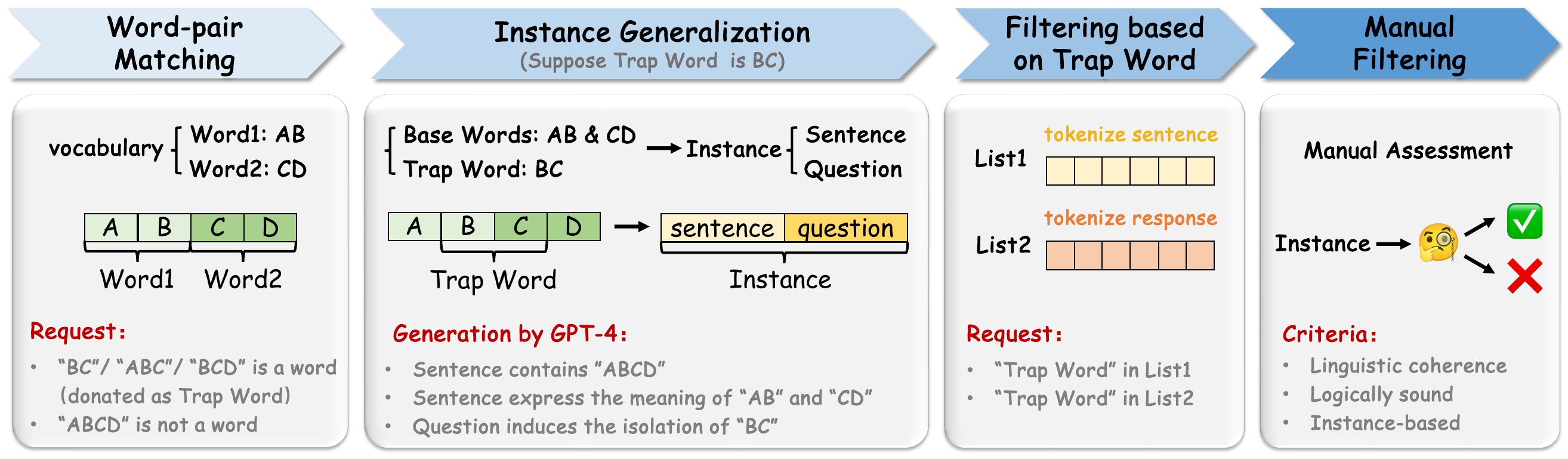}
    \caption{Our framework of generating ADT-Auto automatically.}
    \label{fig:automatically}
\end{figure*}

Given the inefficiency of constructing dataset manually, we further develop an automatic generation framework for dataset to challenge LLMs' tokenization.
As discussed in Section~\ref{Construction}, the challenges of tokenization in English are less severe than in Chinese. 
Consequently, we primarily concentrate on the automatic generation of Chinese data. 
Figure \ref{fig:automatically} illustrates the process of automatically constructing our dataset ADT-Auto.

\subsubsection{Word-pair Matching}
The automatic generation of dataset is also based on exported vocabularies. 
From the vocabularies, we first seek some qualified word pairs.
Given two words \texttt{Word 1} and \texttt{Word 2}, they are considered to be match when meeting the following criteria: 
The suffix (or whole) of \texttt{Word 1} can be concatenated with the prefix (or whole) of \texttt{Word 2}, as a token existing in the vocabulary, denoted as \texttt{Trap Word}. 
Meanwhile, the concatenation of \texttt{Word 1} and \texttt{Word 2} should not be a token existing in the vocabulary.

Accordingly, there are also three situations related to \texttt{Word 1}, \texttt{Word 2} and \texttt{Trap Word}, corresponding to three scenarios in Table \ref{tab:three_methods}.

\begin{enumerate}[topsep=3pt, leftmargin=12pt]
    \setlength{\itemsep}{2pt} 
    \setlength{\parsep}{0pt}  
    \setlength{\parskip}{0pt} 
    
    \item If \texttt{Word 1} is included by \texttt{Trap Word}, this situation corresponds to the schema inserting a span after \texttt{Trap Word}.
    
    \item If \texttt{Word 2} is included by \texttt{Trap Word}, this situation corresponds to the schema inserting a span before \texttt{Trap Word}.
    
    \item If neither \texttt{Word 1} nor \texttt{Word 2} is included by \texttt{Trap Word}, this situation corresponds to the schema inserting a span both before and after \texttt{Trap Word}.
\end{enumerate}

Above clarifications indicate that the concatenation of \texttt{Word 1} and \texttt{Word 2} (as `ABCD' in Figure \ref{fig:automatically}) corresponds to the challenging span in the construction process of ADT-Human. 
Our goal is to compose a sentence that not only convey the semantic essence of \texttt{Word 1} and \texttt{Word 2}, but also induce LLM to isolate \texttt{Trap Word} when tokenization. 
Notably, we ignore the situation that the concatenation of \texttt{Word 1} and \texttt{Word 2} is just \texttt{Trap Word}, as it would not pose challenges to LLMs' tokenization.

To augment the matching efficacy of \texttt{Trap Word} upon matching, we also consider the criterion that the remaining parts of \texttt{Word 1} and \texttt{Word 2} excluded by \texttt{Trap Word} should also exist in the vocabulary as a token. 
Furthermore, the first or last character of \texttt{Trap Word} cannot be a Chinese stop character.

\subsubsection{Instance Generalization} 
With each \texttt{Word 1}, \texttt{Word 2} and their corresponding \texttt{Trap Word}, we harness GPT-4 to generate an instance of ADT-Auto, which consists of a sentence and a question. 
Specifically, we prompt GPT-4 to generate a sentence that is used to challenge the tokenization of LLMs. 
To this end, the generated sentence is required to include the concatenation of \texttt{Word 1} and \texttt{Word 2} (inevitably including \texttt{Trap Word}). 
In addition, GPT-4 is also required to devise a question related to the generated sentence, which is used to evaluate LLMs' tokenization performance through their answers. 
The instance must meet criteria as below: The sentence should convey the meanings of both \texttt{Word 1} and \texttt{Word 2}, while the question's answer should come from \texttt{Word 1}, \texttt{Word 2}, \texttt{Trap Word} or their combination. 
Thus, the influence of LLMs' incorrect tokenization can be identified by checking answers to the question. 

The prompt for GPT-4 to generate instances includes some demonstration examples in addition to task instruction, which is illustrated in Appendix~\ref{app:prompt}.

\subsubsection{Filtering Based on Trap Word}
\label{sec:filter1}
The goal of our dataset is to expose the flaw of LLMs' tokenization. 
Therefore, we will only retain the instances that can induce tokenization problem of LLMs, so we check each instance generated at the previous step. 
For these instances, the presence of \texttt{Trap Word} implies a challenging case that is likely to induce tokenization problems. 
Given an instance, we retain it if its corresponding \texttt{Trap Word} is both in the LLM's tokenization list for the instance's sentence (as List 1 in Figure \ref{fig:automatically}) and in its tokenization list for the answer (as List 2 in Figure \ref{fig:automatically}). 
Such filtering criterion indicates that the LLM commits tokenization errors on understanding the sentence and response for the instance. 

\subsubsection{Manual Filtering}
\label{sec:filter2}
To ensure the retained instances can induce tokenization problems and meanwhile have reasonable expressions, we further adopt manual assessment for instances. 
Specifically, we select the high-qualify instances with considering sentence expressions and LLM's responses. 
Notably, we might still retain some instances to which the used LLM has correct response, since the other LLMs are still likely to commit inaccurate tokenization for these instances, resulting in unsatisfactory responses.

Due to space limitation, we take Qwen-7B as an example to illustrate the process of generating instances. 
There are 24,953 Chinese tokens in Qwen-7B's vocabulary, and after word-pair matching, 1,764,692 word-pairs are obtained. 
From the matching word-pairs, 8,000 pairs are selected randomly and used for instance generation by GPT-4. 
Due to the inherent stochastic characteristic of LLMs on response generation, we conduct three iterations of filtering operations introduced in Subsection \ref{sec:filter1} to get more qualified instances, and thus retain 894 instances. 
Next, through the manual filtering introduced in Subsection \ref{sec:filter2}, we retain 231 instances for ADT-Auto in the end.



\section{Experiments}
\label{Experiments}


\subsection{Experiment Setup}

Considering the open-source LLMs used in our experiments, we select the LLMs previously used in the construction of ADT-Human, including Chatglm3-6B, Baichuan2-13B-Chat, Yi-34B-Chat, Qwen-7B-Chat, and Qwen1.5-72B-Chat for Chinese data, as well as Llama-3-8B-Instruct, Llama-3-70B-Instruct, and Mixtral-8x7B-Instruct-v0.1 for English data. 
We test these LLMs using both locally deployed versions and API versions, with the exception of Chatglm3-6B, which does not have API version.
For the closed-source LLMs, we test GPT-4o, GPT-4, GPT-3.5-Turbo, Qwen2.5-max, step-1-8k\footnote{\url{https://platform.stepfun.com/docs/llm/text}}, moonshot-v1-8k\footnote{\url{https://platform.moonshot.cn/docs}}, ERNIE-3.5-8K\footnote{ \url{https://cloud.baidu.com/doc/WENXINWORKSHOP/s/jlil56u11}} for Chinese data, and GPT-4o, GPT-4 and GPT-3.5-Turbo for English data.
Additionally, we test the API version of Deepseek-R1, which has recently gained significant attention, for Chinese data.

In our experiments, we directly use the dataset ADT constructed with the method introduced in Section \ref{Method}, which includes the manually constructed ADT-Human (containing Chinese and English instances) and the automatically generated ADT-Auto (only containing Chinese instances).

In addition, we conduct our experiments on the platform with four A800 GPUs.

\subsection{ADT-Human's Challenges to LLMs}
\label{experiment_human}

Firstly, we investigate the challenges posed by the manually constructed dataset ADT-Human to LLMs. 
Specifically, we evaluate LLMs' performance through counting the number of incorrect answers generated by LLMs for the questions in instances.
Recalling the process of manually composing challenging spans introduced in Section \ref{Construction}, the span `BC' in Table \ref{tab:three_methods} is in fact the \texttt{Trap Word} mentioned in Section \ref{AutoGen}.
Hence, for a given LLM, its answer including a \texttt{Trap Word} is identified as inaccurate response for the question undoubtedly.
We identify the correctness of LLMs' responses through human assessment.

The percentage of incorrect responses provided by the LLMs
for Chinese data (corresponding to the four LLMs' vocabularies) and English data (corresponding to the two LLMs' vocabularies) are presented in Table~\ref{tab:ADT_Chinese} and Table~\ref{tab:ADT_English}, respectively. 
The results show that ADT-Human poses a significant challenge to both open-source and closed-source LLMs on their tokenization, resulting in very high rates of inaccurate responses. 
It is also worth mentioning that the recent state-of-the-art (SOTA) GPT-4o cannot outperform GPT-4 , implying that the advancements of these LLMs have not yet addressed this primary but challenging problem.

\begin{table*}
  \centering
  \begin{adjustbox}{max width=0.92\textwidth}
  \begin{tabular}{llccccc}
    \toprule
    & \multirow{2}{*}{\textbf{Model}} & \multicolumn{4}{c}{\textbf{Source LLM of vocabulary}} & \multirow{2}{*}{\textbf{Overall error rate}}\\
     & & Chatglm3 & Baichuan2 & Yi & Qwen\\
    \midrule
    \multirow{5}{*}{\textbf{Open-source (Local)}} & Chatglm3-6B & 100.00 & 100.00 & 100.00 & 90.91 & \textbf{97.73} \\
     & Baichuan2-13B-Chat & 90.91 & 100.00 & 81.82 & 100.00 & \textbf{93.18} \\
     & Yi-34B-Chat & 72.73 & 63.64 & 100.00 & 100.00 & \textbf{84.09} \\
     & Qwen-7B-Chat & 100.00 & 72.73 & 90.91 & 100.00 & \textbf{90.91} \\
     & Qwen1.5-72B-Chat & 90.91 & 45.45 & 81.82 & 100.00 & \textbf{79.55} \\
     \midrule
    \multirow{5}{*}{\textbf{Open-source (API)}} & Baichuan2-13B-Chat & 100.00 & 100.00 & 90.91 & 100.00 & \textbf{97.73} \\
     & Yi-34B-Chat & 81.82 & 54.55 & 100.00 & 90.91 & \textbf{81.82} \\
     & Qwen-7B-Chat & 100.00 & 81.82 & 81.82 & 100.00 & \textbf{90.91} \\
     & Qwen1.5-72B-Chat & 90.91 & 54.55 & 81.82 & 100.00 & \textbf{81.82} \\
     & DeepSeek-R1 & 36.36 & 27.27 & 45.45 & 54.55 & \textbf{40.91}\\
    \midrule
    \multirow{7}{*}{\textbf{Closed-source}} & GPT-4o & 72.73 & 27.27 & 54.55 & 90.91 & \textbf{61.36} \\
     & GPT-4 & 81.82 & 45.45 & 45.45 & 27.28 & \textbf{50.00} \\
     & GPT-3.5-Turbo & 72.73 & 27.27 & 72.73 & 72.73 & \textbf{61.36} \\
     & Qwen2.5-max & 90.91 & 72.73 & 90.91 & 100.00 & \textbf{88.64}\\
     & step-1-8k & 63.64 & 18.18  & 72.73 & 63.64 & \textbf{54.55} \\
     & moonshot-v1-8k & 81.82 & 27.27 & 100.00 & 81.82 & \textbf{72.73} \\
     & ERNIE-3.5-8K & 72.73 & 54.55 & 54.55 & 72.73 & \textbf{63.64} \\
    \bottomrule
  \end{tabular}
  \end{adjustbox}
  \caption{Error rates (\%) of answers on ADT-Human (Chinese).}
  \label{tab:ADT_Chinese}
\end{table*}

\begin{table*}
  \centering
  \begin{adjustbox}{max width=0.88\textwidth}
  \begin{tabular}{llccc}
    \toprule
    & \multirow{2}{*}{\textbf{Model}} & \multicolumn{2}{c}{\textbf{Source LLM of vocabulary}} & \multirow{2}{*}{\textbf{Overall error rate}}\\
     & & Llama-3 & Mixtral\\
    \midrule
    \multirow{3}{*}{\textbf{Open-source (Local)}} & Llama-3-8B-Instruct & 100.00 & 85.71 & \textbf{92.86} \\
     & Llama-3-70B-Instruct & 57.14 & 71.43 & \textbf{64.29} \\
     & Mixtral-8x7B-Instruct-v0.1 & 85.71 & 100.00 & \textbf{92.86} \\
    \midrule
    \multirow{3}{*}{\textbf{Open-source (API)}} & Llama-3-8B-Instruct & 100.00 & 71.43 & \textbf{85.71} \\
     & Llama-3-70B-Instruct & 57.14 & 28.57 & \textbf{42.86} \\
     & Mixtral-8x7B-Instruct-v0.1 & 71.43 & 100.00 & \textbf{85.71} \\
    \midrule
    \multirow{3}{*}{\textbf{Closed-source}} & GPT-4o & 57.14 & 71.43 & \textbf{64.29} \\
     & GPT-4 & 57.14 & 57.14 & \textbf{57.14} \\
     & GPT-3.5-Turbo & 71.43 & 57.14 & \textbf{64.29} \\
    \bottomrule
  \end{tabular}
  \end{adjustbox}
  \caption{Error rates (\%) of answers on ADT-Human (English).}
  \label{tab:ADT_English}
\end{table*}

To further investigate the impact of tokenization on the performance of LLMs, a fine-grained analysis of the relationship between tokenization and response is conducted. 
In Appendix~\ref{App: the relationship between tokenization and response}, we quantitatively examine the relationship between the correctness of tokenization and the correctness of LLM's response. 
Given the inaccessibility of tokenization lists for closed-source LLMs, we perform a statistical analysis on open-source LLMs. 
Since LLMs' API versions do not directly provide tokenization lists, the tokenization results obtained from the corresponding locally deployed versions are used. 
Appendix~\ref{App: the relationship between tokenization and response in ADT-Human(Chinese)} and Appendix~\ref{App: the relationship between tokenization and response in ADT-Human(English)} respectively illustrate the quantitative relationships between tokenization and response for the LLMs tested on Chinese and English data of ADT-Human.

\begin{figure}[t]
    \centering
    \begin{minipage}[c]{0.6\columnwidth}
        \centering
        \resizebox{\textwidth}{!}{
        \begin{tabular}{c|cc}
             & Right & Wrong\\
             & response & response \\
            \hline
            Right tokenization & 3 & 3 \\
            Wrong tokenization & 1 & 37 \\
        \end{tabular}}
    \end{minipage}
    \hfill
    \begin{minipage}[c]{0.37\columnwidth} 
        \centering
        \includegraphics[width=\textwidth]{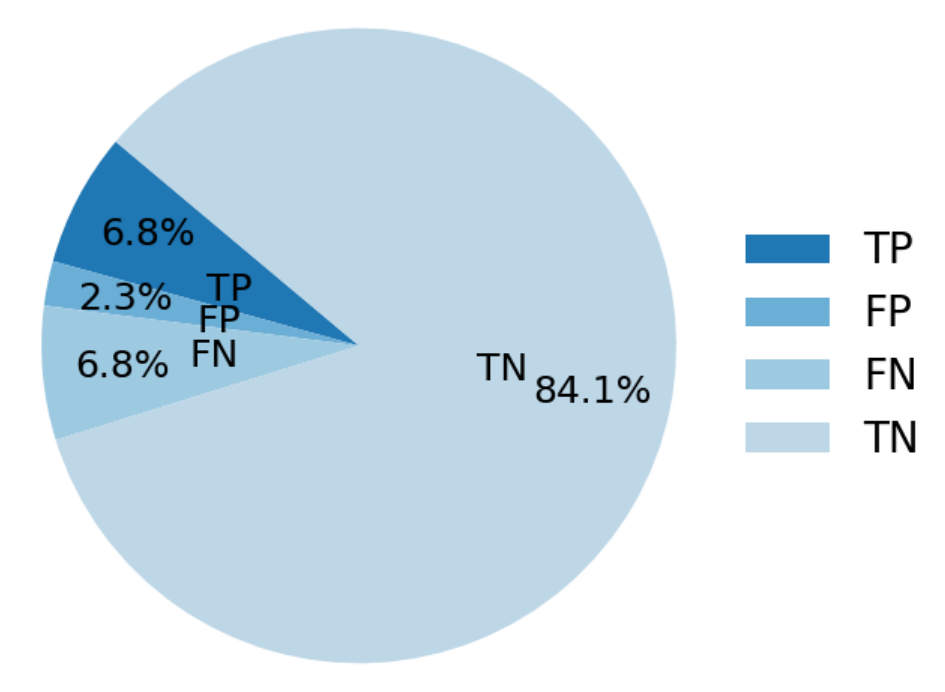}
    \end{minipage}
    \caption{Four relationships between tokenization and response, take Qwen-7B-Chat as an example.}
    \label{fig:quantitative analysis}
\end{figure}

This study further intuitively illustrates the relationships between tokenization and response in the form of pie charts in Appendix~\ref{App:Proportion of four situations between tokenization and response}. In Appendix~\ref{App: Proportion of four situations between tokenization and response in ADT-Human(Chinese)} and Appendix~\ref{App: Proportion of four situations between tokenization and response in ADT-Human(English)}, the proportions of four relationships between tokenization and response for each LLM on Chinese and English data of ADT-Human are displayed respectively. 
Figure~\ref{fig:quantitative analysis} illustrates the quantitative relationships and pie chart using Qwen-7B-Chat as an example. For more details, please refer to Appendix~\ref{App: the relationship between tokenization and response} and Appendix~\ref{App:Proportion of four situations between tokenization and response}.
We mainly focus on the proportion of TN (tokenization incorrect and response incorrect).
As shown in pie charts, the proportion of TN cases is very high in ADT-Human, with an average of 80.91\% for Chinese data and 79.78\% for English data. This indicates that tokenization errors significantly affect the accuracy of LLM responses and also demonstrates that ADT-Human effectively challenges the performance of LLMs.

\subsection{ADT-Auto's Challenges to LLMs}

Similar to the investigation of ADT-Human's challenges to LLMs' tokenization, we also evaluate the LLMs' performance on ADT-Auto. 

\begin{table}
  \centering
  \begin{adjustbox}{max width=\columnwidth}
  \begin{tabular}{llcc}
    \toprule
    & \textbf{Model} & \textbf{Fraction} & \textbf{Error rate (\%)}\\
    \midrule
    \multirow{5}{*}{\textbf{Open-source (Local)}} & Chatglm3-6B & 156/231 & \textbf{67.53}\\
     & Baichuan2-13B-Chat & 147/231 & \textbf{63.64} \\
     & Yi-34B-Chat & 90/231 & \textbf{38.96} \\
     & Qwen-7B-Chat & 185/231 & \textbf{80.09} \\
     & Qwen1.5-72B-Chat & 93/231 & \textbf{40.26} \\
    \midrule
    \multirow{5}{*}{\textbf{Open-source (API)}} & Baichuan2-13B-Chat & 167/231 & \textbf{72.29} \\
     & Yi-34B-Chat & 80/231 & \textbf{34.63} \\
     & Qwen-7B-Chat & 160/231 & \textbf{69.26}\\
     & Qwen1.5-72B-Chat & 97/231 & \textbf{41.99} \\
     & DeepSeek-R1 & 57/231 & \textbf{24.68}\\
    \midrule
    \multirow{7}{*}{\textbf{Closed-source}} & GPT-4o & 75/231 & \textbf{32.47} \\
     & GPT-4 & 89/231 & \textbf{38.53} \\
     & GPT-3.5-Turbo & 99/231 & \textbf{42.86} \\
     & Qwen2.5-max & 75/231 & \textbf{32.47} \\
     & step-1-8k & 60/231 & \textbf{25.97} \\
     & moonshot-v1-8k & 60/231 & \textbf{25.97} \\
     & ERNIE-3.5-8K & 51/231 & \textbf{22.08} \\
    \bottomrule
  \end{tabular}
  \end{adjustbox}
  \caption{Error rates of answers on ADT-Auto.}
  \label{tab:The efficacy of Automatic Generated Data}
\end{table}

ADT-Auto's instances come from Qwen-7B's vocabulary, and the rates of LLMs' inaccurate responses for these instances' questions are listed in Table~\ref{tab:The efficacy of Automatic Generated Data}. 
Based on the results, we have the following observations and analysis:

\begin{enumerate}[topsep=3pt, leftmargin=12pt]
    \setlength{\itemsep}{2pt} 
    \setlength{\parsep}{0pt}  
    \setlength{\parskip}{0pt} 
    \item Compared with closed-source LLMs, open-source LLMs suffer from ADT-Auto's challenges more apparently. 
    It implies that these closed-source LLMs, as the profit-making flagships of their creator companies, naturally have stronger capabilities than open-source LLMs that are created for public usage.

    \item Compared with ADT-Human, ADT-Auto is less challenging to LLMs, since the sentences generated by GPT-4 have more formal, regular or simple syntaxes and expressions than the manually composed sentences in ADT-Human. 
    Thus, these sentences in ADT-Auto are relatively easy for LLMs' understanding.
     
    \item Qwen1.5-72B-Chat has lower error rates than Qwen-7B-Chat, although they have the same vocabulary. 
    We specially check some instances to which Qwen1.5-72B-Chat gives correct answers but Qwen-7B-Chat gives wrong answers, and find the two models have the same incorrect tokenization lists for these instances. 
    It suggests that their different performance is not caused by tokenization. 
    The results also imply that in the case of incorrect tokenization, the bigger LLMs are more robust and likely to generate correct responses than the smaller LLMs thanks to their stronger capabilities brought by the larger scale.
\end{enumerate}

Similar to Section~\ref{experiment_human}, Appendix~\ref{App: the relationship between tokenization and response in ADT-Auto} presents the quantitative relationships between tokenization and response for each open-source LLM tested on ADT-Auto. The corresponding pie charts are shown in Appendix~\ref{App: Proportion of four situations between tokenization and response in ADT-Auto}.
As indicated by the pie charts, the proportion of TN cases is also very high on ADT-Auto, with an average of 46.11\%. 
This further demonstrates that tokenization errors significantly impact the accuracy of LLMs' responses and highlights the effectiveness of ADT-Auto in challenging the performance of LLMs.

\section{Conclusion}
\label{Conclusion}


In this paper, we dedicate to deeply investigating the relationship between LLMs' vulnerability on tokenization and their unsatisfactory responses for some tasks. 
To this end, we construct an adversarial dataset \textbf{ADT (Adversarial Dataset for Tokenizer)} consisting of a manually constructed subset ADT-Human and an automatically generated subset ADT-Auto, of which each instance includes a sentence and a corresponding question. 
Our experiments demonstrate that our dataset does challenge the studied open-source and closed-source LLMs' token segmentation, resulting in their incorrect answers. 
We hope our work and dataset could shed light on the subsequent research on improving LLMs' performance.

\section*{Limitations}

The key contribution of this study lies in drawing attention to the impact of tokenization on LLM's performance and providing two frameworks for data generation.  
As for how to propose improvement strategies based on this phenomenon, we are currently in the process of exploration.
Additionally, this study focuses primarily on Chinese. Whether other languages are similarly affected by tokenization remains to be further investigated.

\section*{Ethics Statement}
\label{Ethics Statement}

We hereby declare that all authors of this article are aware of and adhere to the provided ACL Code of Ethics and honor the code of conduct.
\paragraph*{Use of Human Annotations.}
Human annotations are only used in methodological research at the beginning of the work, to assist in analyzing the feasibility of the proposed solution. Annotators consented to the use of data for research purposes. We ensure the privacy of all annotators is protected throughout the annotation process, and all of them are adequately paid according to local standards.
\paragraph*{Risks.}
Synthetic data generated by LLMs may involve potential ethical risks regarding fairness and bias \cite{bommasani2021opportunities, blodgett-etal-2020-language}, which results in further consideration when they are employed in downstream tasks. We asked our members for proofreading to refine the offensive and harmful data generated by GPT-4. Despite these considerations, there may still be some unsatisfactory data that goes unnoticed in our final dataset.

\bibliography{anthology,custom}

\appendix

\newpage

\section{Details of ADT-Human}
\label{app:ADT}

\vspace{1em} 

\subsection{Chatglm3}

\begin{figure}[h]
    \centering
    \includegraphics[width=1\linewidth]{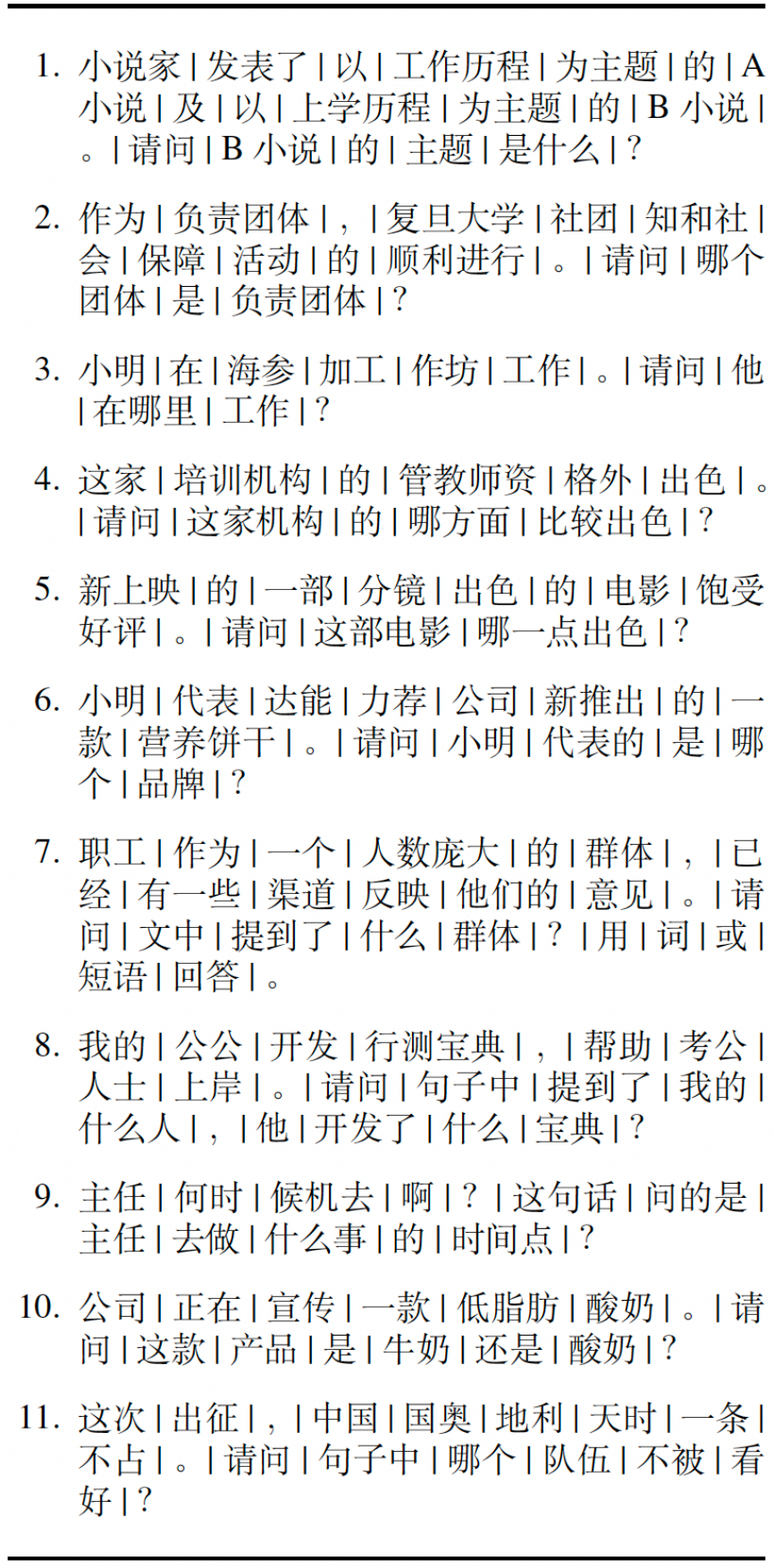}
\end{figure}

\newpage
\subsection{Baichuan2}

\vspace{1em} 

\begin{figure}[h]
    \centering
    \includegraphics[width=1\linewidth]{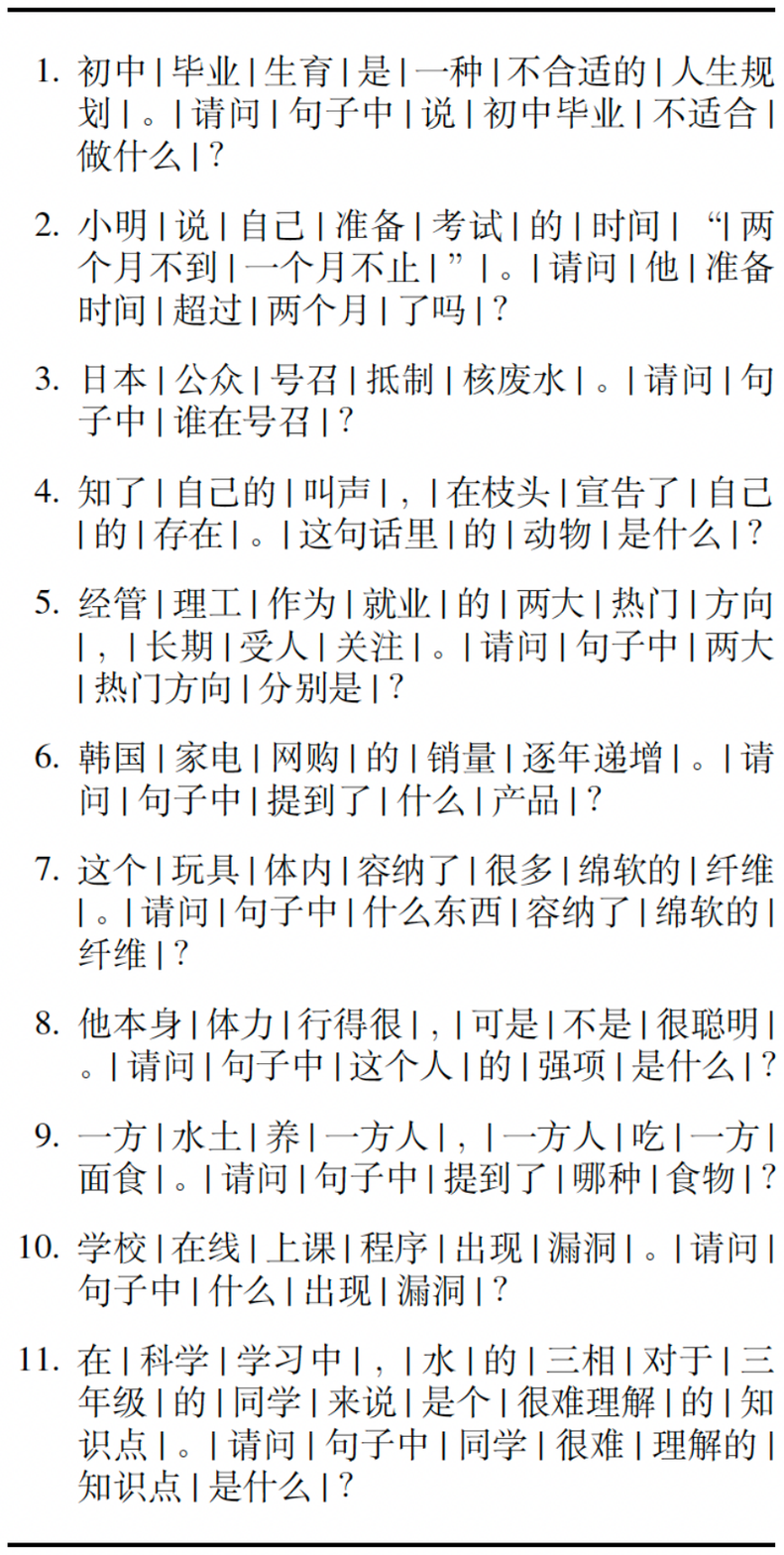}
\end{figure}

\newpage
\subsection{Yi}

\vspace{1em} 

\begin{figure}[H]
    \centering
    \includegraphics[width=1\linewidth]{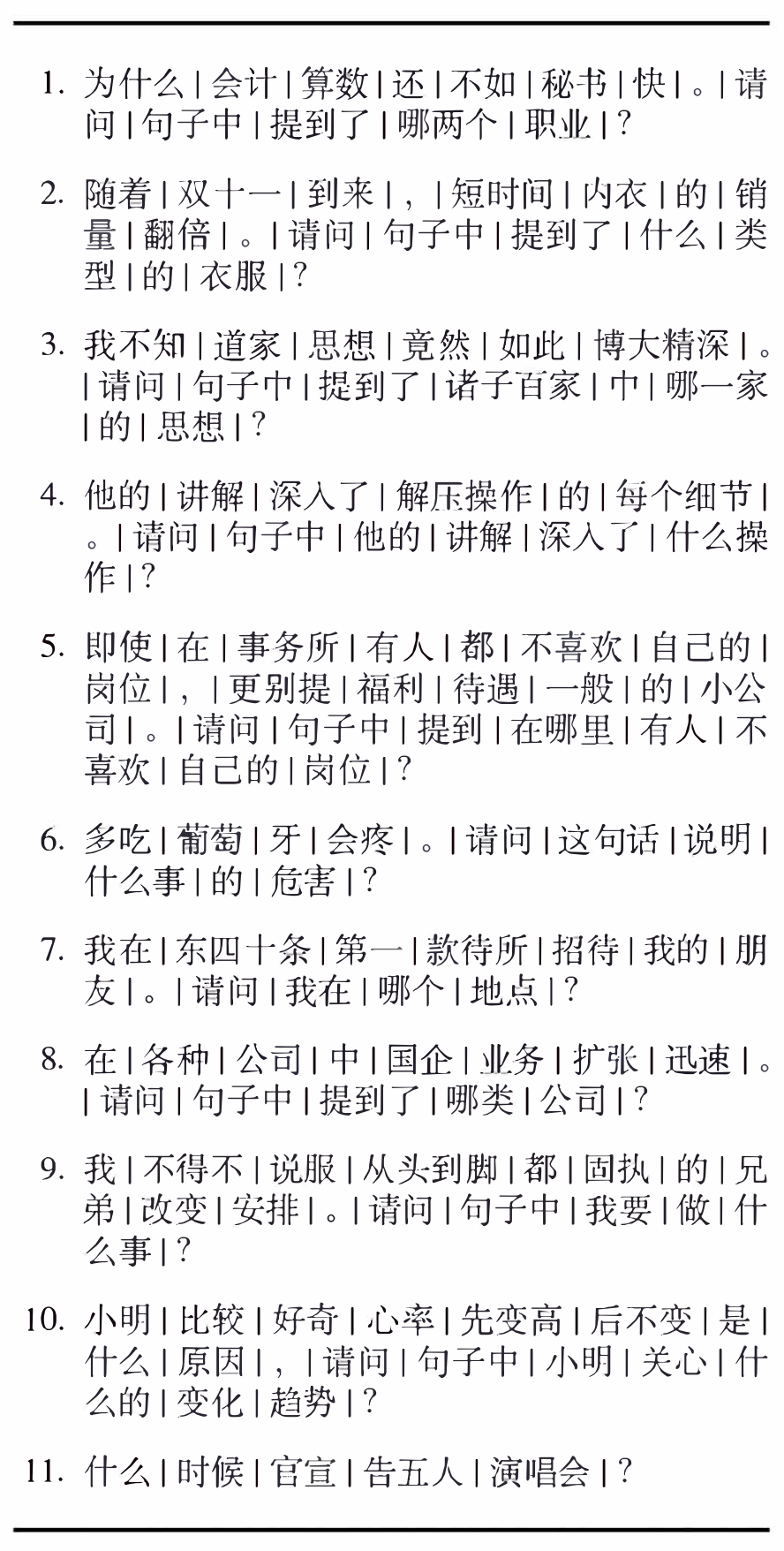}
    \label{fig:yi}
\end{figure}

\newpage
\subsection{Qwen}

\vspace{1em} 

\begin{figure}[H]
    \centering
    \includegraphics[width=1\linewidth]{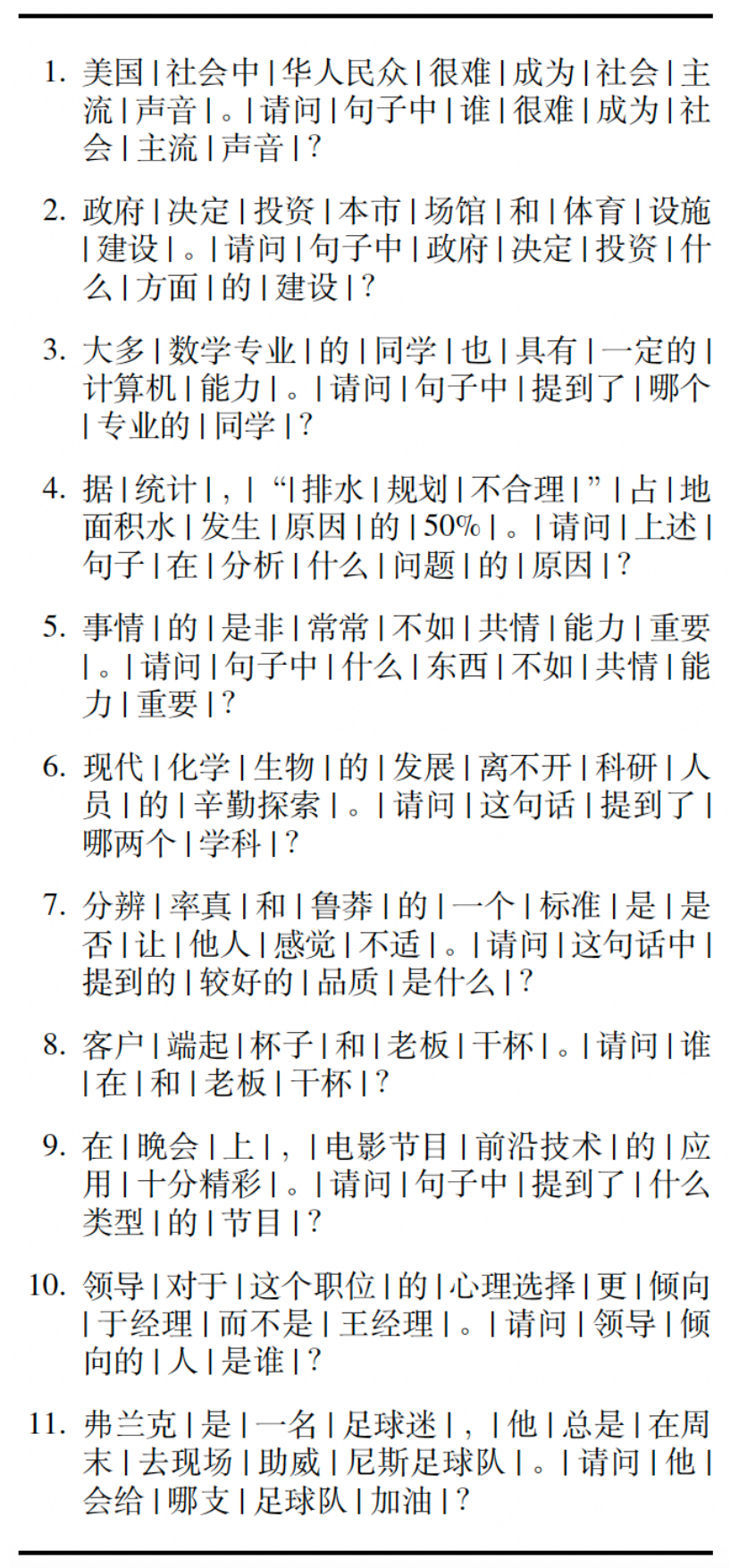}
    \label{fig:qwen}
\end{figure}

\newpage
\subsection{Llama-3}

\begin{minipage}{7.7cm}
\rule{\linewidth}{1pt}
\vspace{-0.3cm}
{\setstretch{0.9}
\begin{enumerate}
    \item In the postal history course, today's homework is to \textbf{list|envelope|classes} in history. What should I do?
    \item As a Marathon lover, in the past he only \textbf{ran|domestically}, but now he also goes abroad. What behavior of him is discussed in the sentence?
    \item As fans of Kenshi Yonezu, they \textbf{sing|lemon} in the auditorium. What song are they singing in the sentence?
    \item  I wanted to go to the beach, \textbf{but|tons|of} work piled up. How much work did I have?
    \item  We are measuring how \textbf{fat|her|cat} is to make sure her cat is healthy. What metric are we measuring?
    \item This automotive shop mainly sells \textbf{car|bonnets}. What does the shop sell?
    \item The researcher was disappointed to \textbf{miss|ionization} in the sample, which was crucial for the experiment's success. What made the researcher disappointed?
\end{enumerate}}
\vspace{-0.5cm}
\rule{\linewidth}{1pt}
\end{minipage}

\subsection{Mixtral}

\begin{minipage}{7.7cm}
\rule{\linewidth}{1pt}
\vspace{-0.3cm}
{\setstretch{0.9}
\begin{enumerate}
    \item The analyst emphasized the importance of tracking \textbf{live|return} to gauge real-time performance. What importance was emphasized by the analyst?
    \item Many countries \textbf{import|sports} from another country. What do those countries import?
    \item He \textbf{moves|table} to another place. What is he doing?
    \item Those pants and leather shoes \textbf{fitted|speakers} very well. Who do those pants and leather shoes fit well?
    \item The soccer team \textbf{won|derby} against its rival. Did the soccer team win or lose?
    \item They \textbf{swap|pears} with each other. What are they exchanging?
    \item The \textbf{leg|ends|up} being the most injured part, requiring immediate medical attention. In the sentence, which part is injured the most?
\end{enumerate}}
\vspace{-0.5cm}
\rule{\linewidth}{1pt}
\end{minipage}

\section{The relationship between tokenization and response}
\label{App: the relationship between tokenization and response}

\vspace{-4pt}
\subsection{ADT-Human (Chinese)}
\label{App: the relationship between tokenization and response in ADT-Human(Chinese)}

\begin{minipage}{\columnwidth}
\end{minipage}
\vspace{-5.5pt}

\begin{minipage}{\columnwidth}
    \centering
    \begin{tabular}{c|cc}
         & Right & Wrong\\
         & response & response \\
        \hline
        Right tokenization & 2 & 1 \\
        Wrong tokenization & 1 & 40 \\
    \end{tabular}
    \captionsetup{labelformat=empty}
    \captionof{table}{(a) Baichuan2-13B-Chat (Local)}
\end{minipage}

\vspace{5pt}
\begin{minipage}{\columnwidth}
    \centering
    \begin{tabular}{c|cc}
         & Right & Wrong\\
         & response & response \\
        \hline
        Right tokenization & 1 & 2 \\
        Wrong tokenization & 0 & 41 \\
    \end{tabular}
    \captionsetup{labelformat=empty}
    \captionof{table}{(b) Baichuan2-13B-Chat (API)}
\end{minipage}

\vspace{5pt}
\begin{minipage}{\columnwidth}
    \centering
    \begin{tabular}{c|cc}
         & Right & Wrong\\
         & response & response \\
        \hline
        Right tokenization & 3 & 0 \\
        Wrong tokenization & 5 & 36 \\
    \end{tabular}
    \captionsetup{labelformat=empty}
    \captionof{table}{(c) Yi-34B-Chat (Local)}
\end{minipage}

\vspace{5pt}
\begin{minipage}{\columnwidth}
    \centering
    \begin{tabular}{c|cc}
         & Right & Wrong\\
         & response & response \\
        \hline
        Right tokenization & 3 & 0 \\
        Wrong tokenization & 7 & 34 \\
    \end{tabular}
    \captionsetup{labelformat=empty}
    \captionof{table}{(d) Yi-34B-Chat (API)}
\end{minipage}

\vspace{5pt}
\begin{minipage}{\columnwidth}
    \centering
    \begin{tabular}{c|cc}
         & Right & Wrong\\
         & response & response \\
        \hline
        Right tokenization & 3 & 3 \\
        Wrong tokenization & 1 & 37 \\
    \end{tabular}
    \captionsetup{labelformat=empty}
    \captionof{table}{(e) Qwen-7B-Chat (Local)}
\end{minipage}

\vspace{5pt}
\begin{minipage}{\columnwidth}
    \centering
    \begin{tabular}{c|cc}
         & Right & Wrong\\
         & response & response \\
        \hline
        Right tokenization & 3 & 3 \\
        Wrong tokenization & 1 & 37 \\
    \end{tabular}
    \captionsetup{labelformat=empty}
    \captionof{table}{(f) Qwen-7B-Chat (API)}
\end{minipage}

\vspace{5pt}
\begin{minipage}{\columnwidth}
    \centering
    \begin{tabular}{c|cc}
         & Right & Wrong\\
         & response & response \\
        \hline
        Right tokenization & 6 & 0 \\
        Wrong tokenization & 3 & 35 \\
    \end{tabular}
    \captionsetup{labelformat=empty}
    \captionof{table}{(g) Qwen1.5-72B-Chat (Local)}
\end{minipage}

\vspace{5pt}
\begin{minipage}{\columnwidth}
    \centering
    \begin{tabular}{c|cc}
         & Right & Wrong\\
         & response & response \\
        \hline
        Right tokenization & 6 & 0 \\
        Wrong tokenization & 2 & 36 \\
    \end{tabular}
    \captionsetup{labelformat=empty}
    \captionof{table}{(h) Qwen1.5-72B-Chat (API)}
\end{minipage}

\vspace{5pt}
\begin{minipage}{\columnwidth}
    \centering
    \begin{tabular}{c|cc}
         & Right & Wrong\\
         & response & response \\
        \hline
        Right tokenization & 1 & 1 \\
        Wrong tokenization & 0 & 42 \\
    \end{tabular}
    \captionsetup{labelformat=empty}
    \captionof{table}{(i) Chatglm3-6B (Local)}
\end{minipage}

\vspace{5pt}
\begin{minipage}{\columnwidth}
    \centering
    \begin{tabular}{c|cc}
         & Right & Wrong\\
         & response & response \\
        \hline
        Right tokenization & 3 & 0 \\
        Wrong tokenization & 23 & 18 \\
    \end{tabular}
    \captionsetup{labelformat=empty}
    \captionof{table}{(j) Deepseek-R1 (API)}
\end{minipage}

\subsection{ADT-Human (English)}
\label{App: the relationship between tokenization and response in ADT-Human(English)}

\begin{minipage}{\columnwidth}
\end{minipage}
\vspace{-5pt}

\begin{minipage}{\columnwidth}
    \centering
    \begin{tabular}{c|cc}
         & Right & Wrong\\
         & response & response \\
        \hline
        Right tokenization & 1 & 0 \\
        Wrong tokenization & 0 & 13 \\
    \end{tabular}
    \captionsetup{labelformat=empty}
    \captionof{table}{(a) Llama-3-8B-Instruct (Local)}
\end{minipage}

\vspace{5pt}
\begin{minipage}{\columnwidth}
    \centering
    \begin{tabular}{c|cc}
         & Right & Wrong\\
         & response & response \\
        \hline
        Right tokenization & 0 & 1 \\
        Wrong tokenization & 0 & 13 \\
    \end{tabular}
    \captionsetup{labelformat=empty}
    \captionof{table}{(b) Llama-3-8B-Instruct (API)}
\end{minipage}

\vspace{5pt}
\begin{minipage}{\columnwidth}
    \centering
    \begin{tabular}{c|cc}
         & Right & Wrong\\
         & response & response \\
        \hline
        Right tokenization & 0 & 0 \\
        Wrong tokenization & 3 & 11 \\
    \end{tabular}
    \captionsetup{labelformat=empty}
    \captionof{table}{(c) Llama-3-70B-Instruct (Local)}
\end{minipage}

\vspace{5pt}
\begin{minipage}{\columnwidth}
    \centering
    \begin{tabular}{c|cc}
         & Right & Wrong\\
         & response & response \\
        \hline
        Right tokenization & 0 & 0 \\
        Wrong tokenization & 8 & 6 \\
    \end{tabular}
    \captionsetup{labelformat=empty}
    \captionof{table}{(d) Llama-3-70B-Instruct (API)}
\end{minipage}

\vspace{5pt}
\begin{minipage}{\columnwidth}
    \centering
    \begin{tabular}{c|cc}
         & Right & Wrong\\
         & response & response \\
        \hline
        Right tokenization & 0 & 0 \\
        Wrong tokenization & 2 & 12 \\
    \end{tabular}
    \captionsetup{labelformat=empty}
    \captionof{table}{(e) Mixtral-8x7B-Instruct-v0.1 (Local)}
\end{minipage}

\vspace{5pt}
\begin{minipage}{\columnwidth}
    \centering
    \begin{tabular}{c|cc}
         & Right & Wrong\\
         & response & response \\
        \hline
        Right tokenization & 0 & 0 \\
        Wrong tokenization & 2 & 12 \\
    \end{tabular}
    \captionsetup{labelformat=empty}
    \captionof{table}{(f) Mixtral-8x7B-Instruct-v0.1 (API)}
\end{minipage}

\subsection{ADT-Auto}
\label{App: the relationship between tokenization and response in ADT-Auto}

\begin{minipage}{\columnwidth}
\end{minipage}
\vspace{-5pt}

\begin{minipage}{\columnwidth}
    \centering
    \begin{tabular}{c|cc}
         & Right & Wrong\\
         & response & response \\
        \hline
        Right tokenization & 28 & 32 \\
        Wrong tokenization & 56 & 115 \\
    \end{tabular}
    \captionsetup{labelformat=empty}
    \captionof{table}{(a) Baichuan2-13B-Chat (Local)}
\end{minipage}

\vspace{5pt}
\begin{minipage}{\columnwidth}
    \centering
    \begin{tabular}{c|cc}
         & Right & Wrong\\
         & response & response \\
        \hline
        Right tokenization & 23 & 37 \\
        Wrong tokenization & 41 & 130 \\
    \end{tabular}
    \captionsetup{labelformat=empty}
    \captionof{table}{(b) Baichuan2-13B-Chat (API)}
\end{minipage}

\vspace{5pt}
\begin{minipage}{\columnwidth}
    \centering
    \begin{tabular}{c|cc}
         & Right & Wrong\\
         & response & response \\
        \hline
        Right tokenization & 45 & 14 \\
        Wrong tokenization & 96 & 76 \\
    \end{tabular}
    \captionsetup{labelformat=empty}
    \captionof{table}{(c) Yi-34B-Chat (Local)}
\end{minipage}

\vspace{5pt}
\begin{minipage}{\columnwidth}
    \centering
    \begin{tabular}{c|cc}
         & Right & Wrong\\
         & response & response \\
        \hline
        Right tokenization & 48 & 11 \\
        Wrong tokenization & 103 & 69 \\
    \end{tabular}
    \captionsetup{labelformat=empty}
    \captionof{table}{(d) Yi-34B-Chat (API)}
\end{minipage}

\vspace{5pt}
\begin{minipage}{\columnwidth}
    \centering
    \begin{tabular}{c|cc}
         & Right & Wrong\\
         & response & response \\
        \hline
        Right tokenization & 4 & 22 \\
        Wrong tokenization & 42 & 163 \\
    \end{tabular}
    \captionsetup{labelformat=empty}
    \captionof{table}{(e) Qwen-7B-Chat (Local)}
\end{minipage}

\vspace{5pt}
\begin{minipage}{\columnwidth}
    \centering
    \begin{tabular}{c|cc}
         & Right & Wrong\\
         & response & response \\
        \hline
        Right tokenization & 10 & 16 \\
        Wrong tokenization & 61 & 144 \\
    \end{tabular}
    \captionsetup{labelformat=empty}
    \captionof{table}{(f) Qwen-7B-Chat (API)}
\end{minipage}

\vspace{5pt}
\begin{minipage}{\columnwidth}
    \centering
    \begin{tabular}{c|cc}
         & Right & Wrong\\
         & response & response \\
        \hline
        Right tokenization & 21 & 5 \\
        Wrong tokenization & 117 & 88 \\
    \end{tabular}
    \captionsetup{labelformat=empty}
    \captionof{table}{(g) Qwen1.5-72B-Chat (Local)}
\end{minipage}

\vspace{5pt}
\begin{minipage}{\columnwidth}
    \centering
    \begin{tabular}{c|cc}
         & Right & Wrong\\
         & response & response \\
        \hline
        Right tokenization & 20 & 6 \\
        Wrong tokenization & 114 & 91 \\
    \end{tabular}
    \captionsetup{labelformat=empty}
    \captionof{table}{(h) Qwen1.5-72B-Chat (API)}
\end{minipage}

\vspace{5pt}
\begin{minipage}{\columnwidth}
    \centering
    \begin{tabular}{c|cc}
         & Right & Wrong\\
         & response & response \\
        \hline
        Right tokenization & 32 & 21 \\
        Wrong tokenization & 43 & 135 \\
    \end{tabular}
    \captionsetup{labelformat=empty}
    \captionof{table}{(i) Chatglm3-6B (Local)}
\end{minipage}

\vspace{5pt}
\begin{minipage}{\columnwidth}
    \centering
    \begin{tabular}{c|cc}
         & Right & Wrong\\
         & response & response \\
        \hline
        Right tokenization & 59 & 3 \\
        Wrong tokenization & 115 & 54 \\
    \end{tabular}
    \captionsetup{labelformat=empty}
    \captionof{table}{(j) Deepseek-R1 (API)}
\end{minipage}

\section{Proportion of four situations between tokenization and response}
\label{App:Proportion of four situations between tokenization and response}

Define the four relationships between tokenization and response:
\begin{itemize}
    \item \textbf{TP: }Correct tokenization and correct response.
    \item \textbf{FP: }Incorrect tokenization but correct response.
    \item \textbf{FN: }Correct tokenization but incorrect response.
    \item \textbf{TN: }Incorrect tokenization and incorrect response.
\end{itemize}

\subsection{ADT-Human (Chinese)}
\label{App: Proportion of four situations between tokenization and response in ADT-Human(Chinese)}

\begin{figure}[H]
    \centering
    \begin{subfigure}[b]{0.48\columnwidth}
        \includegraphics[width=\textwidth]{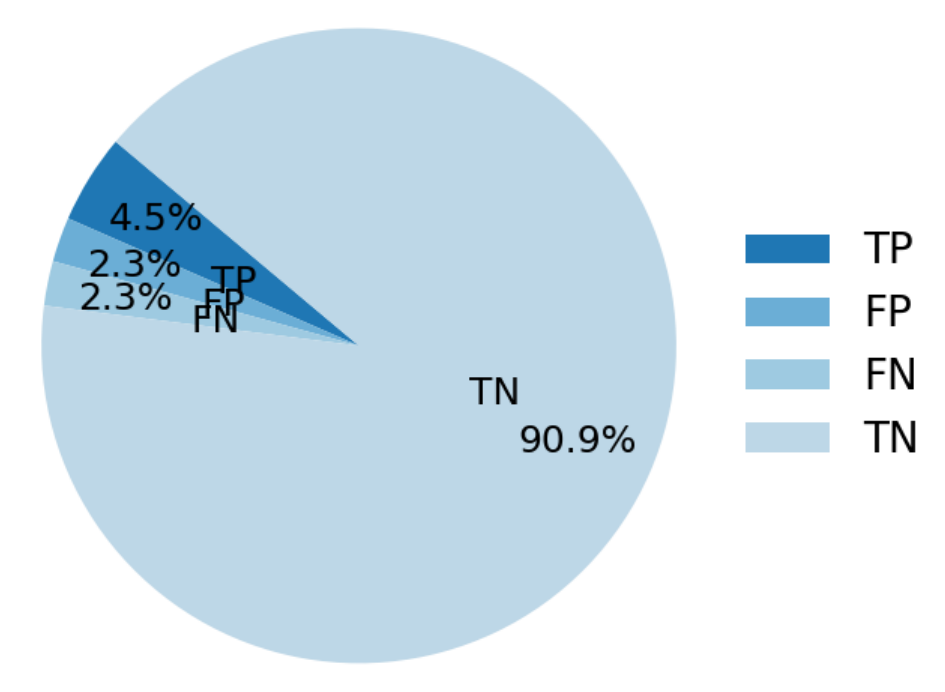}
        \centering
        \caption*{(a) Baichuan2-13B-Chat (Local)}
        \label{fig:Baichuan2-13B-Chat (Local)}
    \end{subfigure}
    \hfill
    \begin{subfigure}[b]{0.48\columnwidth}
        \includegraphics[width=\textwidth]{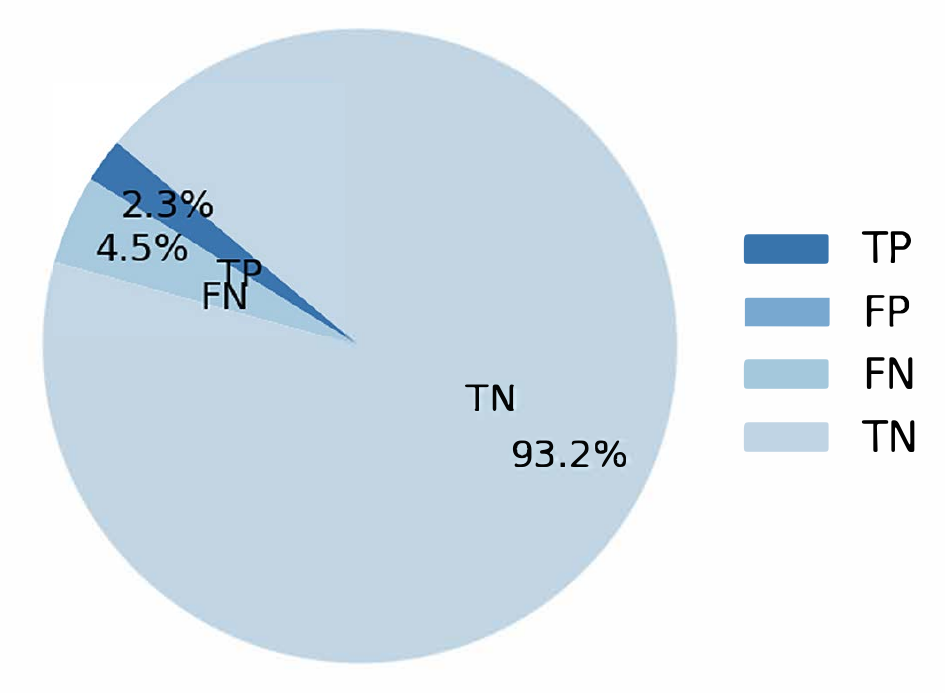}
        \centering
        \caption*{(b) Baichuan2-13B-Chat (API)}
        \label{fig:Baichuan2-13B-Chat (API)}
    \end{subfigure}
\end{figure}

\begin{figure}[H]
    \centering
    \begin{subfigure}[b]{0.48\columnwidth}
        \includegraphics[width=\textwidth]{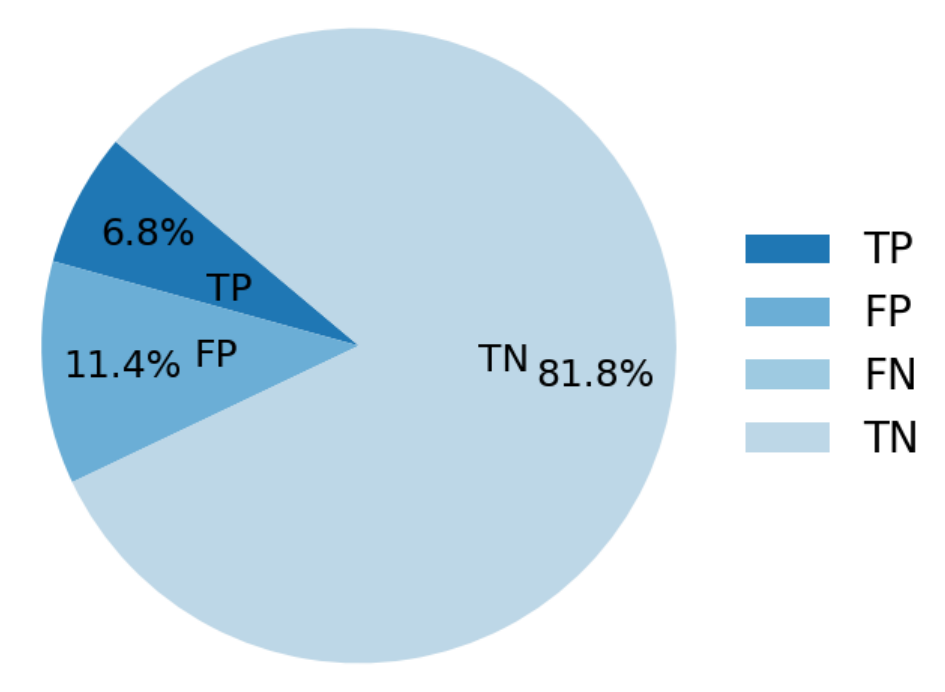}
        \centering
        \caption*{(c) Yi-34B-Chat (Local)}
        \label{fig:Yi-34B-Chat (Local)}
    \end{subfigure}
    \hfill
    \begin{subfigure}[b]{0.48\columnwidth}
        \includegraphics[width=\textwidth]{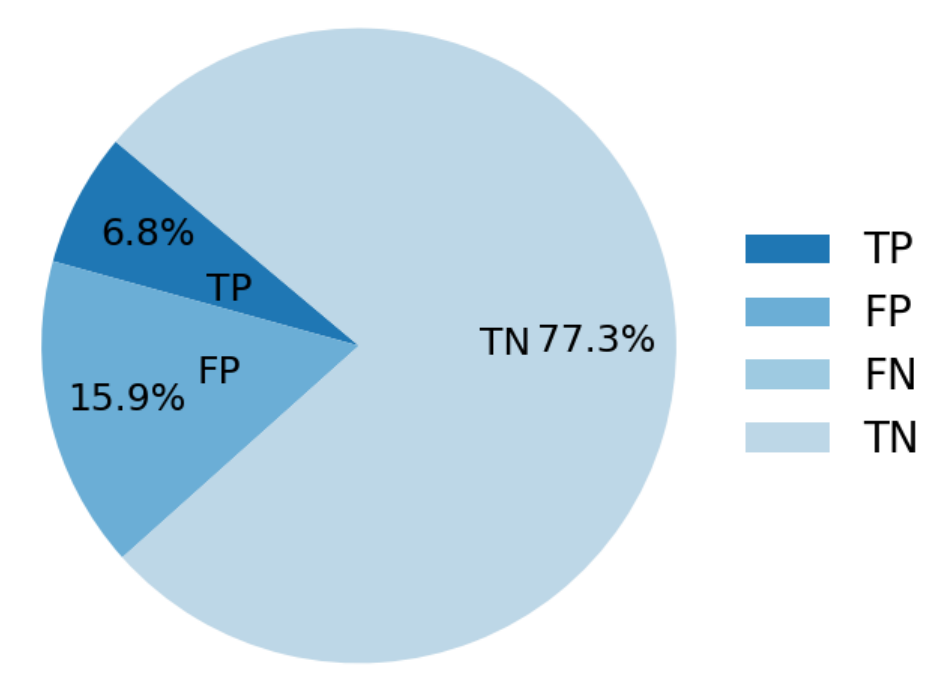}
        \centering
        \caption*{(d) Yi-34B-Chat (API)}
        \label{fig:Yi-34B-Chat (API)}
    \end{subfigure}
\end{figure}

\begin{figure}[H]
    \centering
    \begin{subfigure}[b]{0.48\columnwidth}
        \includegraphics[width=\textwidth]{Figures/pie_chart/Qwen-7B-Chat__Local_.pdf}
        \centering
        \caption*{(e) Qwen-7B-Chat (Local)}
        \label{fig:Qwen-7B-Chat (Local)}
    \end{subfigure}
    \hfill
    \begin{subfigure}[b]{0.48\columnwidth}
        \includegraphics[width=\textwidth]{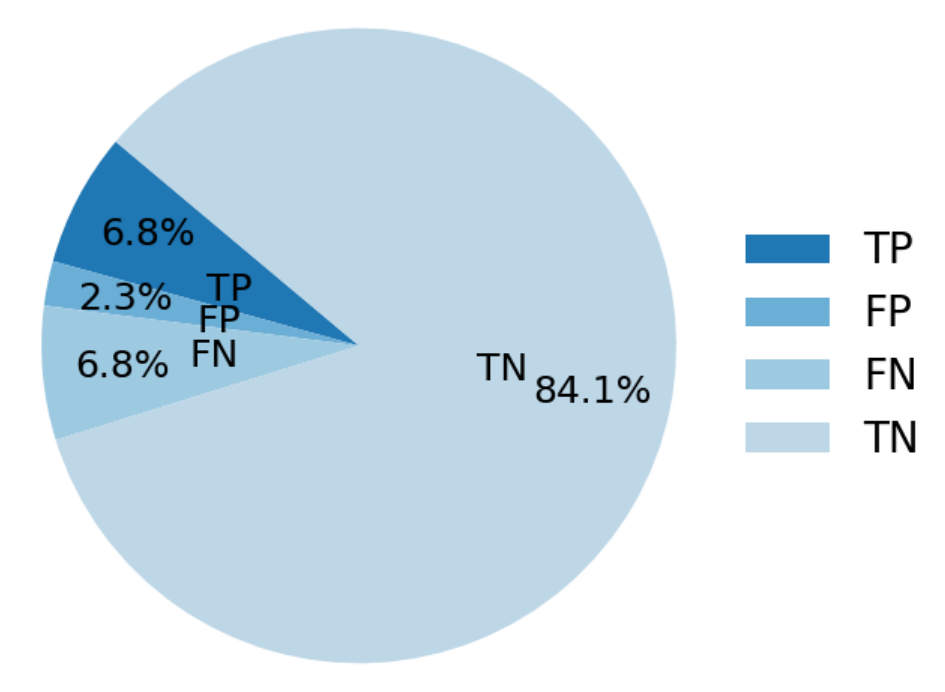}
        \centering
        \caption*{(f) Qwen-7B-Chat (API)}
        \label{fig:Qwen-7B-Chat (API)}
    \end{subfigure}
\end{figure}

\begin{figure}[H]
    \centering
    \begin{subfigure}[b]{0.48\columnwidth}
        \includegraphics[width=\textwidth]{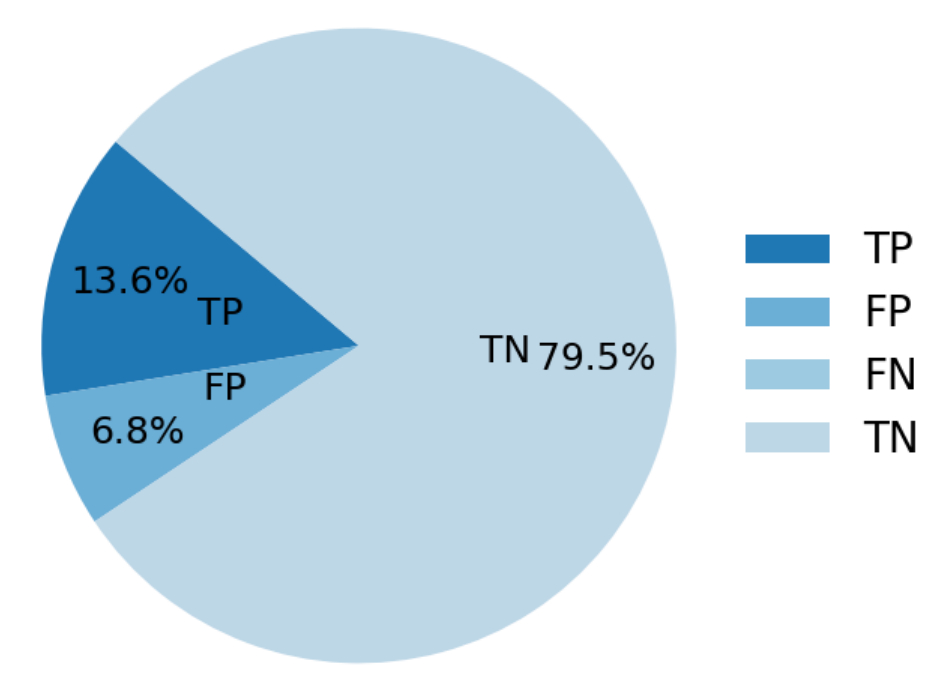}
        \centering
        \caption*{(g) Qwen1.5-72B-Chat (Local)}
        \label{fig:Qwen1.5-72B-Chat (Local)}
    \end{subfigure}
    \hfill
    \begin{subfigure}[b]{0.48\columnwidth}
        \includegraphics[width=\textwidth]{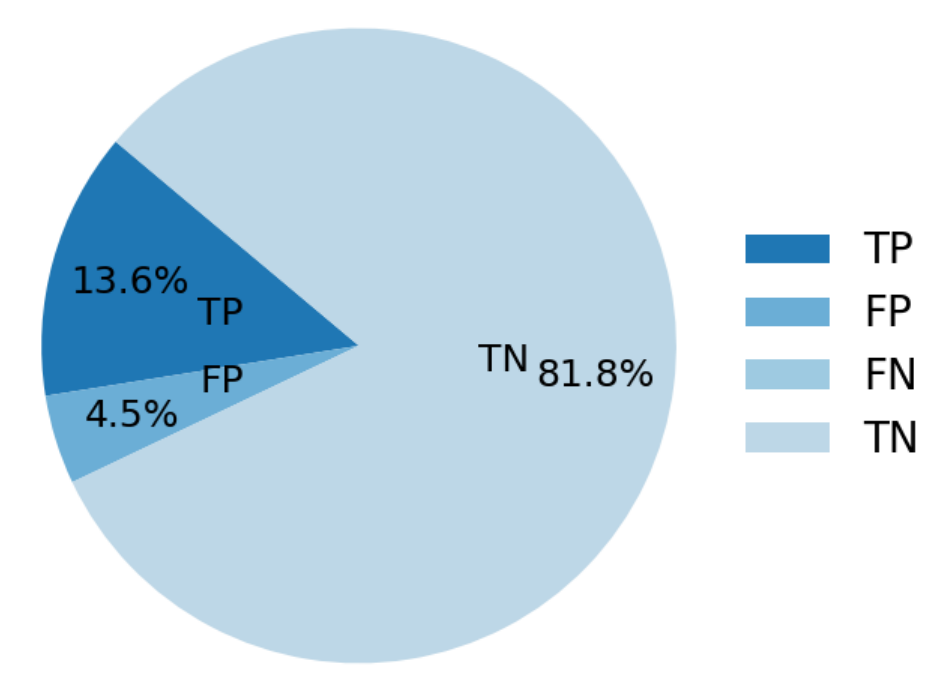}
        \centering
        \caption*{(h) Qwen1.5-72B-Chat (API)}
        \label{fig:Qwen1.5-72B-Chat (API)}
    \end{subfigure}
\end{figure}

\begin{figure}[H]
    \begin{subfigure}[b]{0.48\columnwidth}
        \includegraphics[width=\textwidth]{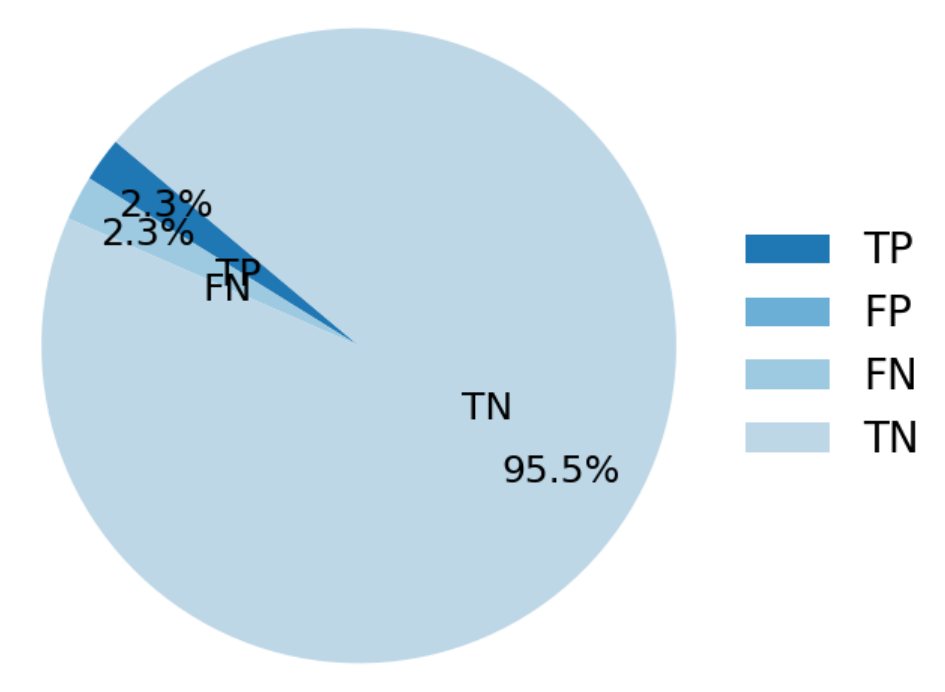}
        \centering
        \caption*{(i) Chatglm3-6B (Local)}
        \label{fig:Chatglm3-6B (Local)}
    \end{subfigure}
    \hfill
    \begin{subfigure}[b]{0.48\columnwidth}
        \includegraphics[width=\textwidth]{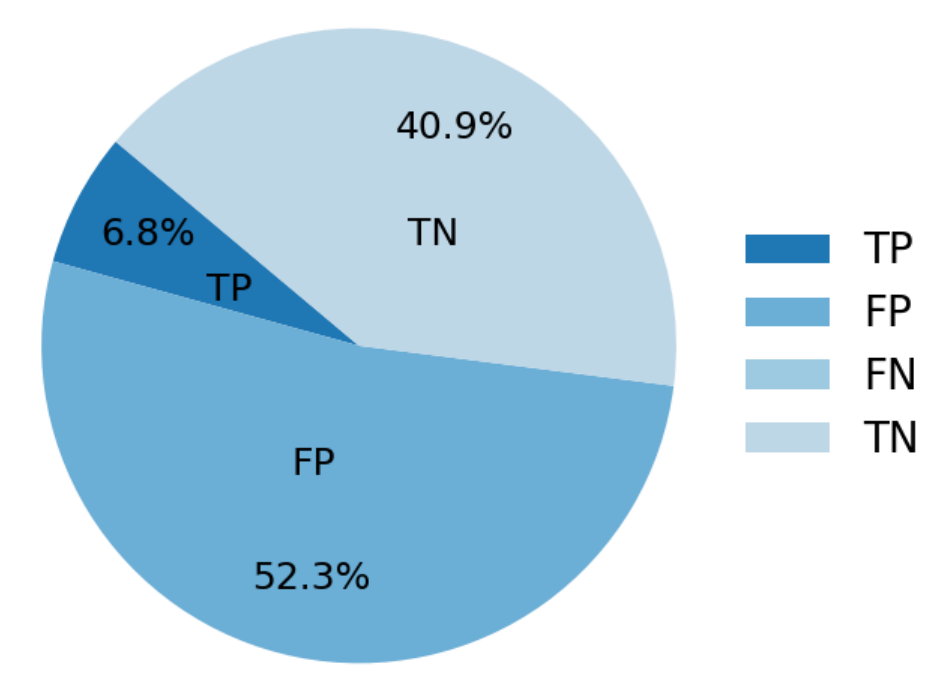}
        \centering
        \caption*{(j) Deepseek-R1 (API)}
        \label{fig:Deepseek-R1 (API)}
    \end{subfigure}
\end{figure}

\subsection{ADT-Human (English)}
\label{App: Proportion of four situations between tokenization and response in ADT-Human(English)}

\begin{figure}[H]
    \centering
    \begin{subfigure}[b]{0.48\columnwidth}
        \includegraphics[width=\textwidth]{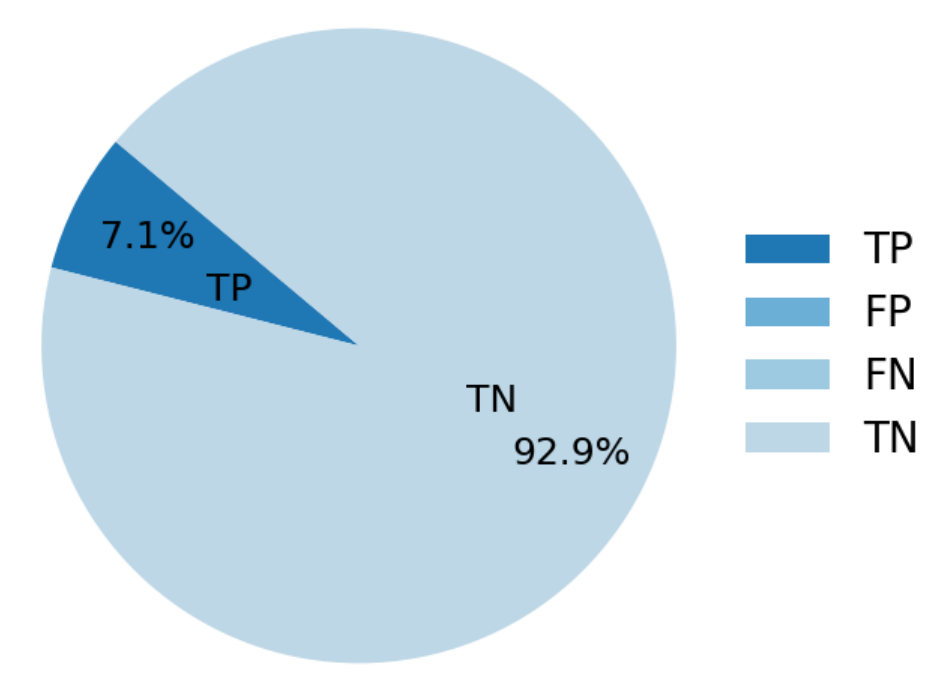}
        \centering
        \caption*{(a) Llama-3-8B-Instruct (Local)}
        \label{fig:Llama-3-8B-Instruct (Local)}
    \end{subfigure}
    \hfill
    \begin{subfigure}[b]{0.48\columnwidth}
        \includegraphics[width=\textwidth]{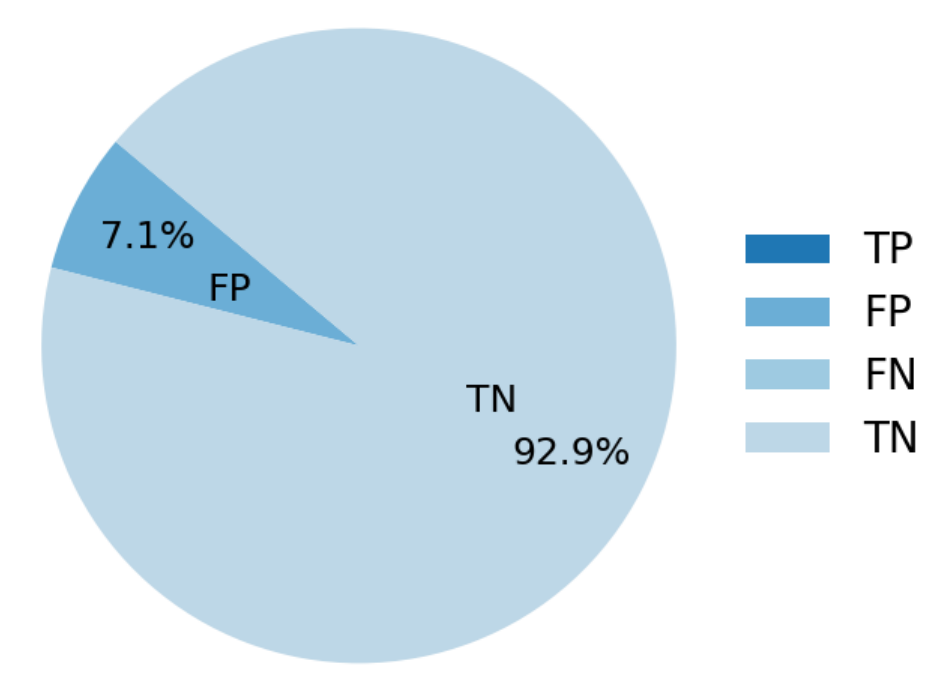}
        \centering
        \caption*{(b) Llama-3-8B-Instruct (API)}
        \label{fig:Llama-3-8B-Instruct (API)}
    \end{subfigure}
\end{figure}

\begin{figure}[H]
    \centering
    \begin{subfigure}[b]{0.48\columnwidth}
        \includegraphics[width=\textwidth]{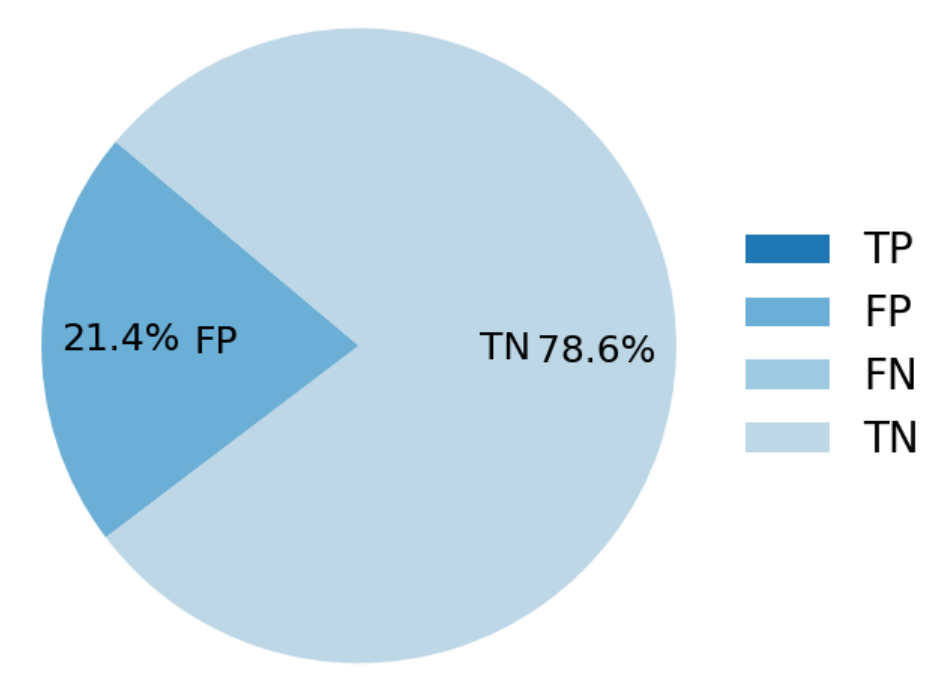}
        \centering
        \caption*{(c) Llama-3-70B-Instruct (Local)}
        \label{fig:Llama-3-70B-Instruct (Local)}
    \end{subfigure}
    \hfill
    \begin{subfigure}[b]{0.48\columnwidth}
        \includegraphics[width=\textwidth]{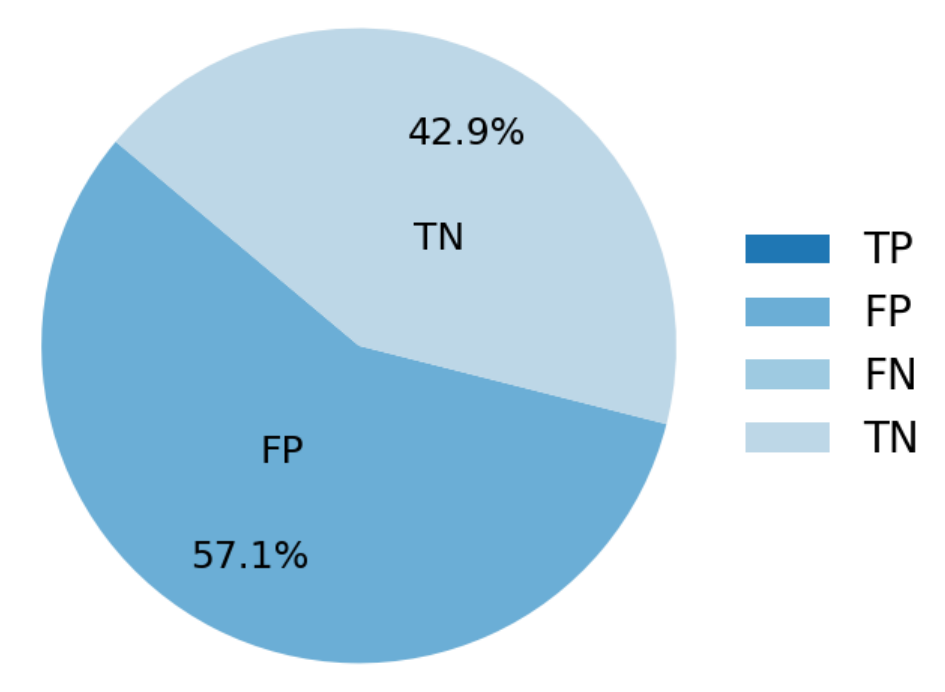}
        \centering
        \caption*{(d) Llama-3-70B-Instruct (API)}
        \label{fig:Llama-3-70B-Instruct(API)}
    \end{subfigure}
\end{figure}

\begin{figure}[H]
    \centering
    \begin{subfigure}[b]{0.48\columnwidth}
        \includegraphics[width=\textwidth]{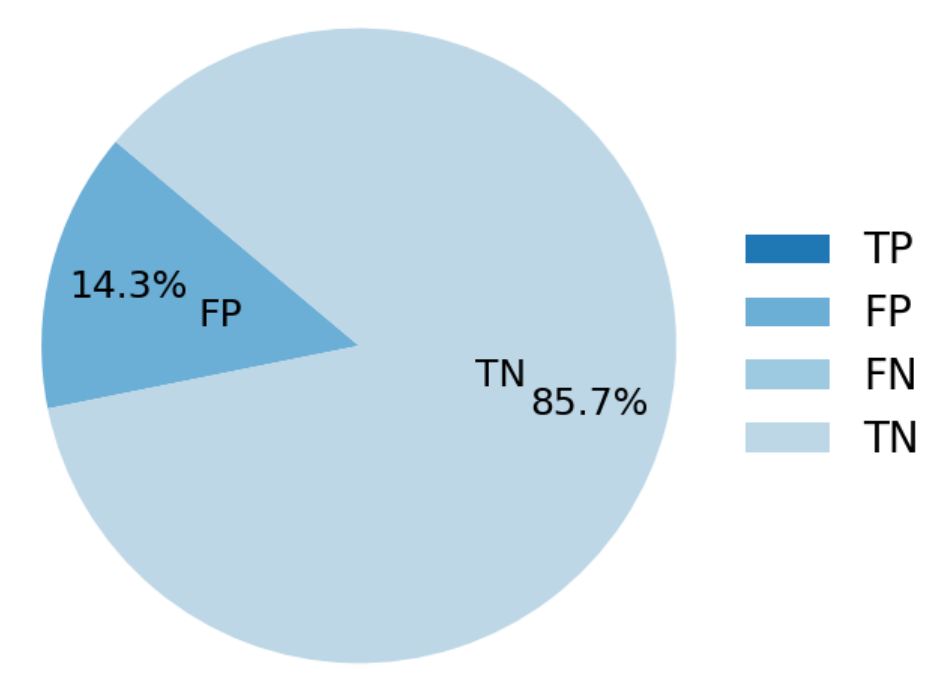}
        \centering
        \caption*{(e) Mixtral-8x7B-Instruct-v0.1 (Local)}
        \label{fig:Mixtral-8x7B-Instruct-v0.1 (Local)}
    \end{subfigure}
    \hfill
    \begin{subfigure}[b]{0.48\columnwidth}
        \includegraphics[width=\textwidth]{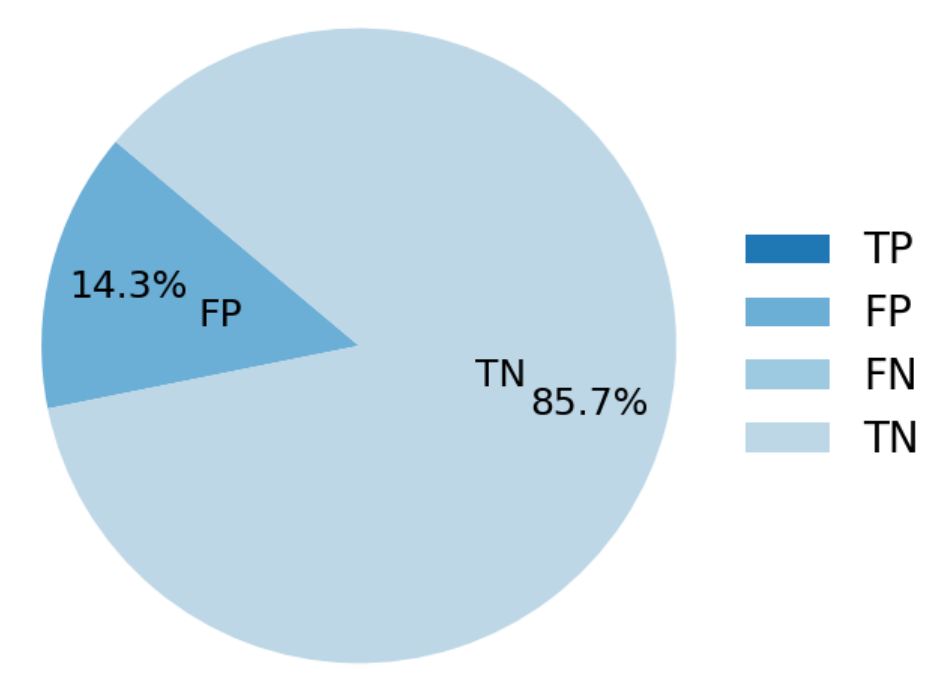}
        \centering
        \caption*{(f) Mixtral-8x7B-Instruct-v0.1 (API)}
        \label{fig:Mixtral-8x7B-Instruct-v0.1 (API)}
    \end{subfigure}
\end{figure}

\subsection{ADT-Auto}
\label{App: Proportion of four situations between tokenization and response in ADT-Auto}

\begin{figure}[!h]
    \centering
    \begin{subfigure}[b]{0.48\columnwidth}
        \includegraphics[width=\textwidth]{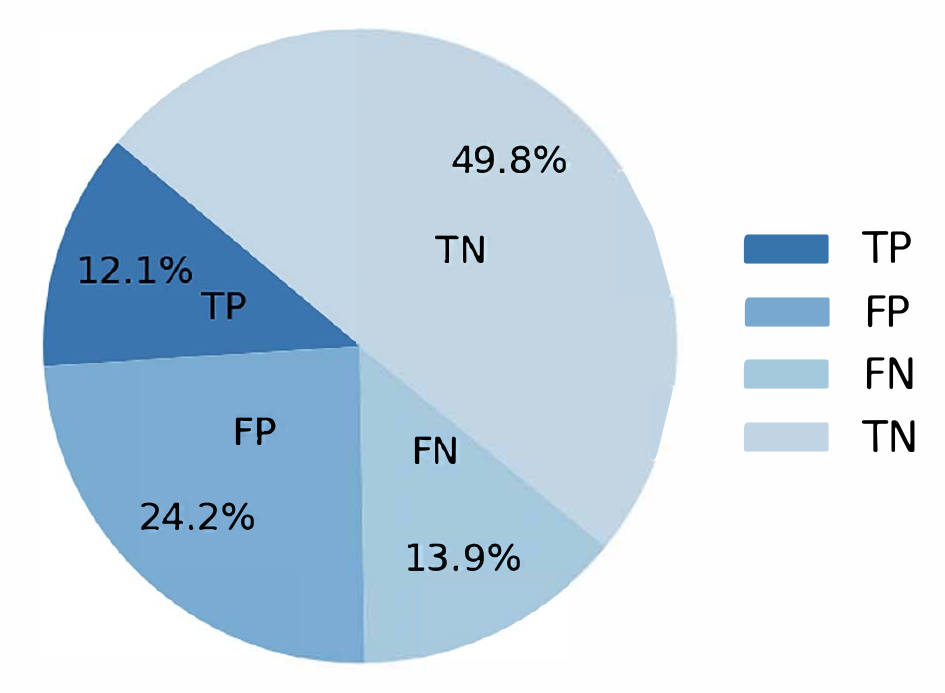}
        \centering
        \caption*{(a) Baichuan2-13B-Chat (Local)}
    \end{subfigure}
    \hfill
    \begin{subfigure}[b]{0.48\columnwidth}
        \includegraphics[width=\textwidth]{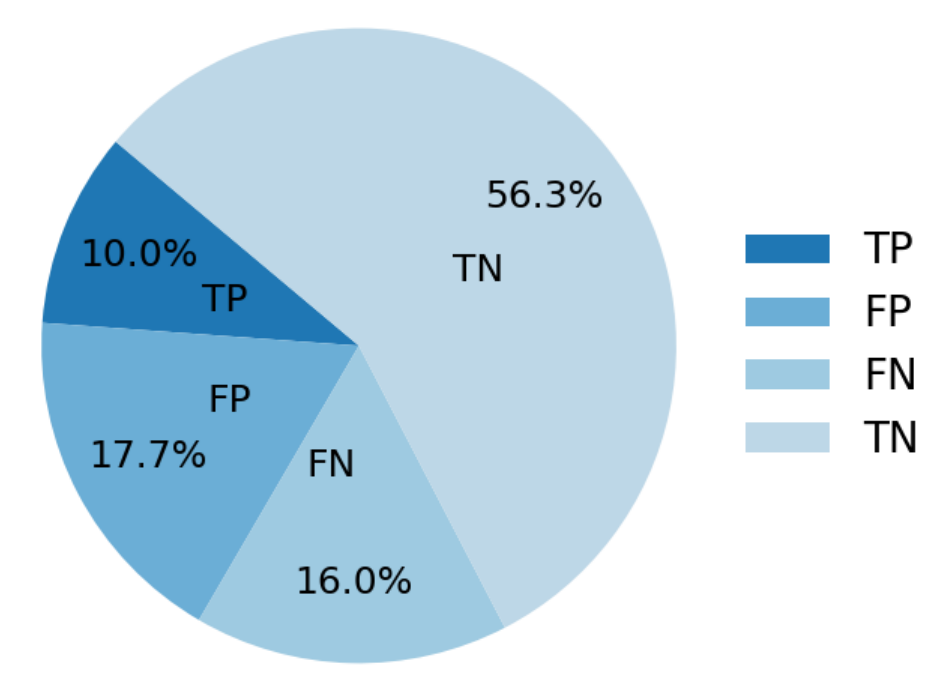}
        \centering
        \caption*{(b) Baichuan2-13B-Chat (API)}
    \end{subfigure}
\end{figure}

\begin{figure}[!h]
    \centering
    \begin{subfigure}[b]{0.48\columnwidth}
        \includegraphics[width=\textwidth]{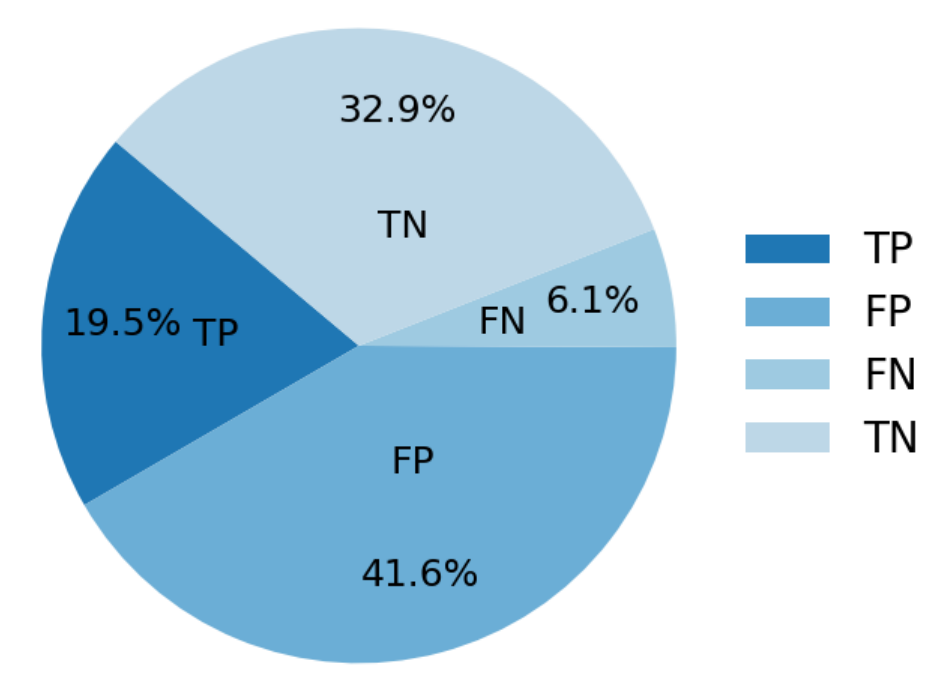}
        \centering
        \caption*{(c) Yi-34B-Chat (Local)}
    \end{subfigure}
    \hfill
    \begin{subfigure}[b]{0.48\columnwidth}
        \includegraphics[width=\textwidth]{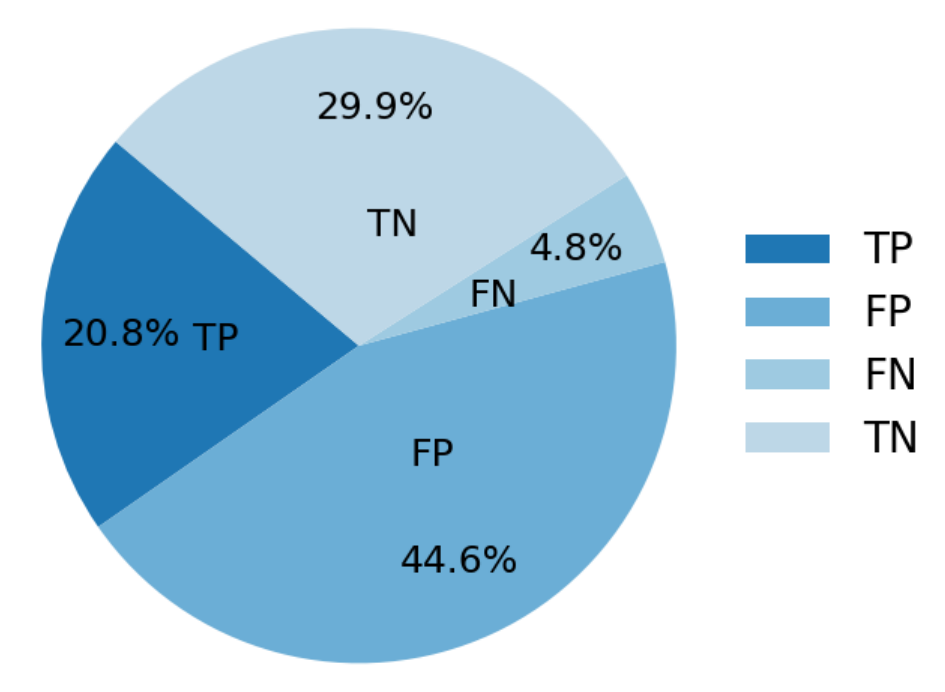}
        \centering
        \caption*{(d) Yi-34B-Chat (API)}
    \end{subfigure}
\end{figure}

\begin{figure}[!h]
    \centering
    \begin{subfigure}[b]{0.48\columnwidth}
        \includegraphics[width=\textwidth]{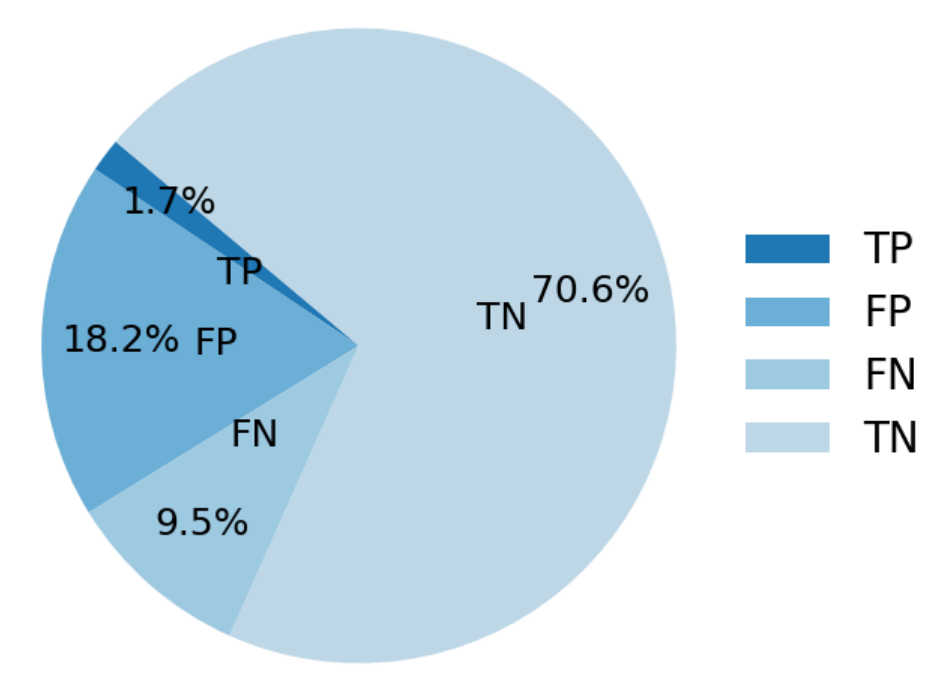}
        \centering
        \caption*{(e) Qwen-7B-Chat (Local)}
    \end{subfigure}
    \hfill
    \begin{subfigure}[b]{0.48\columnwidth}
         \includegraphics[width=\textwidth]{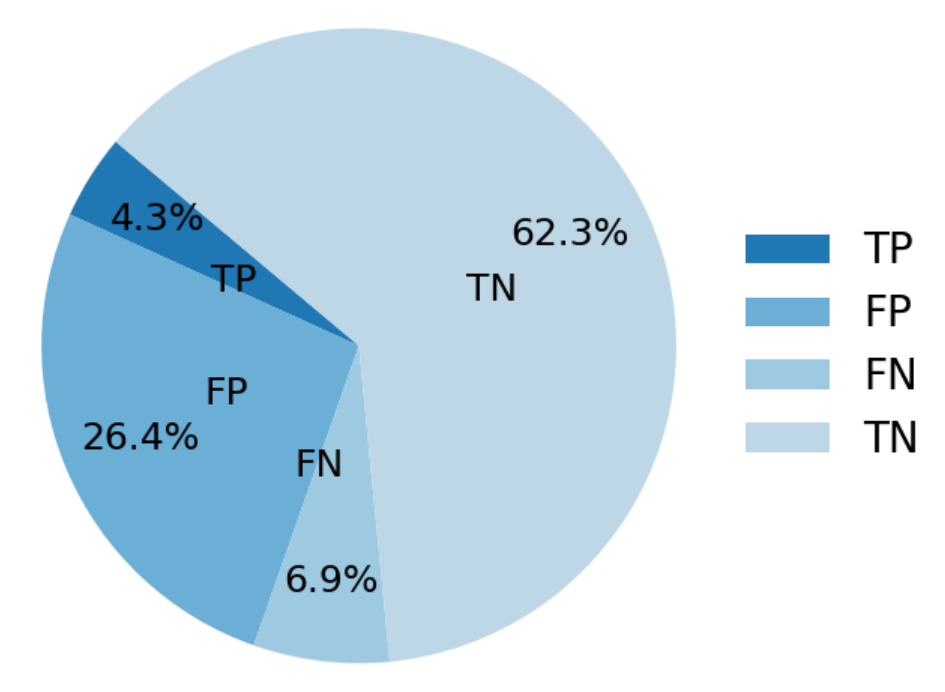}
        \centering
        \caption*{(f) Qwen-7B-Chat (API)}
    \end{subfigure}
\end{figure}

\begin{figure}[!h]
    \centering
    \begin{subfigure}[b]{0.48\columnwidth}
        \includegraphics[width=\textwidth]{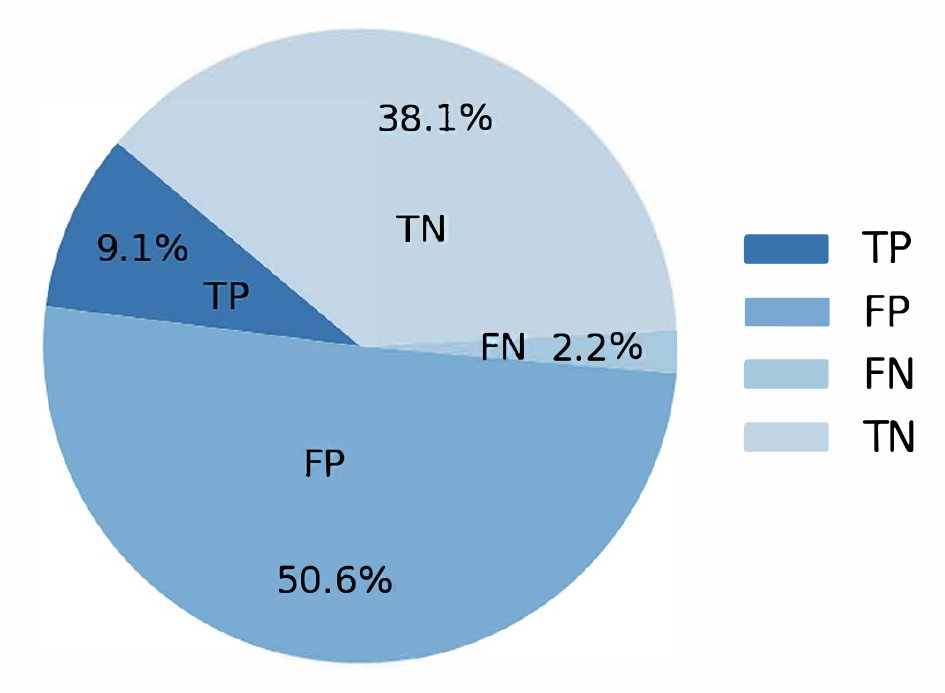}
        \centering
        \caption*{(g) Qwen1.5-72B-Chat (Local)}
    \end{subfigure}
    \hfill
    \begin{subfigure}[b]{0.48\columnwidth}
         \includegraphics[width=\textwidth]{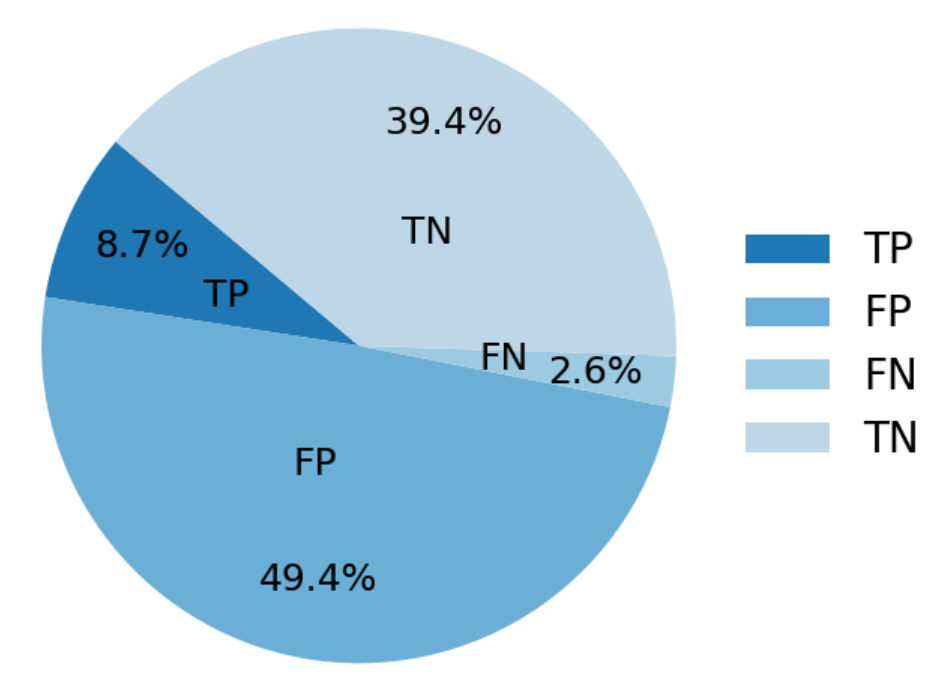}
        \centering
        \caption*{(h) Qwen1.5-72B-Chat (API)}
    \end{subfigure}
\end{figure}

\begin{figure}[H]
    \begin{subfigure}[b]{0.48\columnwidth}
        \includegraphics[width=\textwidth]{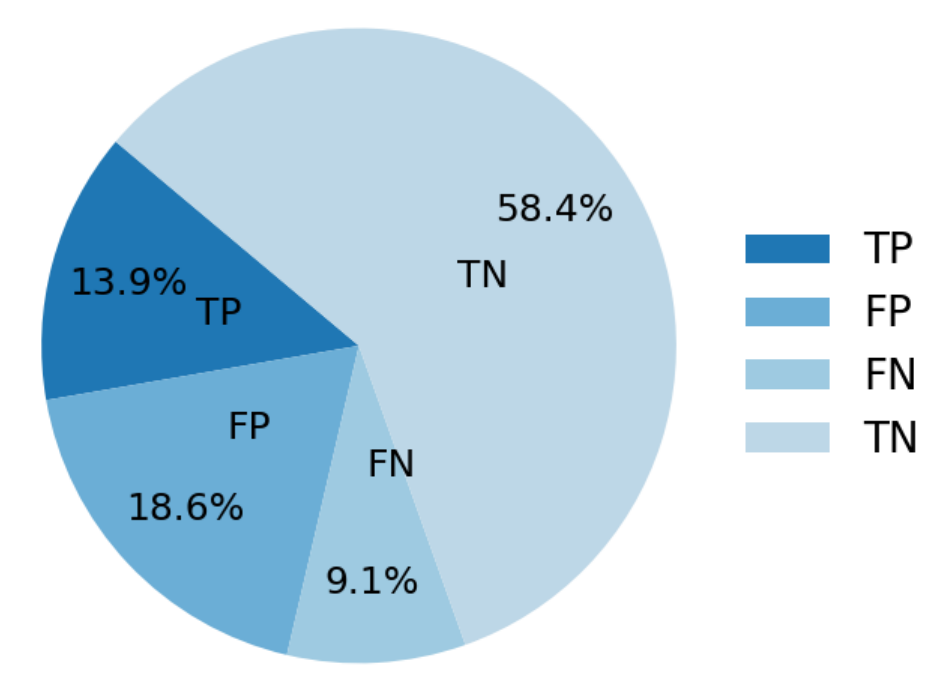}
        \centering
        \caption*{(i) Chatglm3-6B (Local)}
    \end{subfigure}
    \hfill
    \begin{subfigure}[b]{0.48\columnwidth}
         \includegraphics[width=\textwidth]{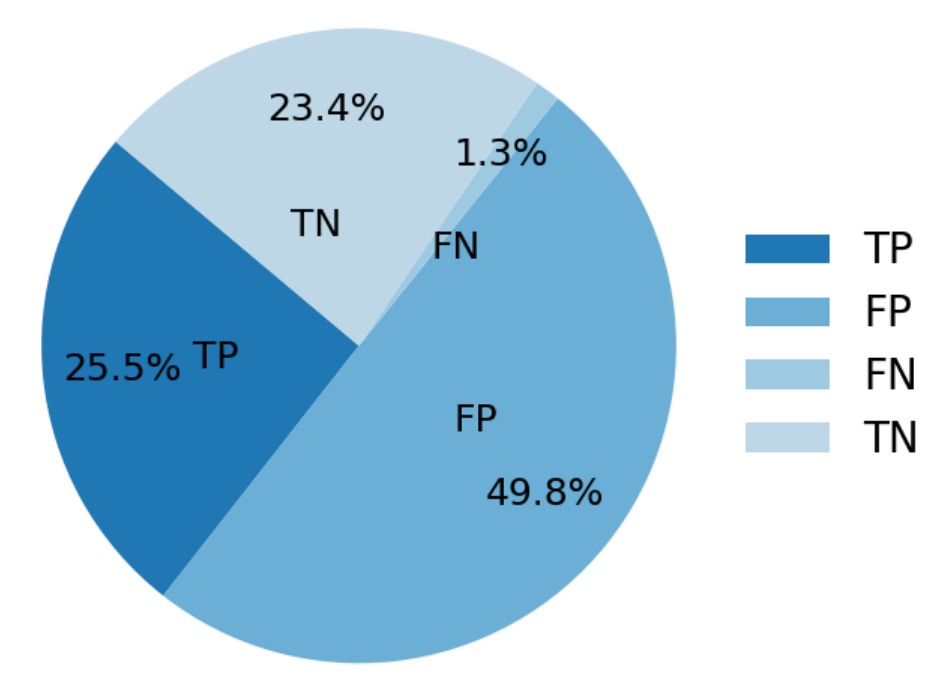}
        \centering
        \caption*{(j) Deepseek-R1 (API)}
    \end{subfigure}
\end{figure}


\section{Details of prompt}
\label{app:prompt}

\begin{figure*}[h]
    \centering
    \includegraphics[width=1\linewidth]{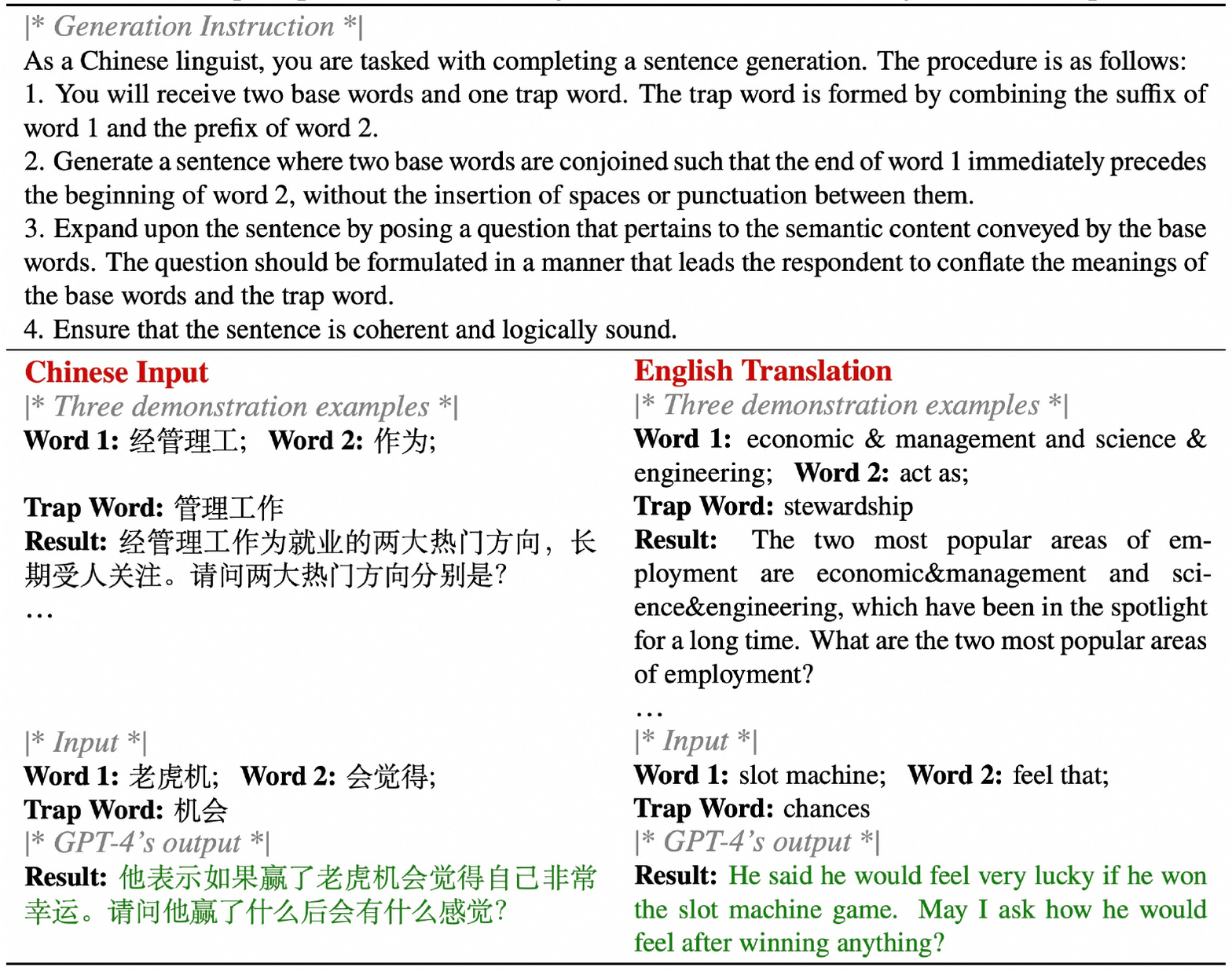}
    \label{fig:prompt}
\end{figure*}

\end{document}